\documentclass{article}
\usepackage{comment,bm,subcaption,float,graphicx,amsfonts,amsmath,amssymb,amsthm}

\usepackage{fullpage}

\usepackage[numbers]{natbib}

\usepackage{hyperref}

\usepackage[inline]{enumitem}

\usepackage[capitalise,nameinlink]{cleveref}

\newcommand{\ww}{\widehat{\bm w}}
\newcommand{\norm}[1]{\left\lVert#1\right\rVert_2}
\newcommand{\smallnorm}[1]{\lVert#1\rVert}
\newcommand{\gl}{\bm g_{\operatorname{lead}}}
\newcommand{\glp}{\bm g'_{\operatorname{lead}}}
\newcommand{\gr}{\bm g_{\operatorname{rest}}}
\newcommand{\grp}{\bm g'_{\operatorname{rest}}}
\newcommand{\esp}{\widehat{\bm \Sigma}'}
\newcommand{\es}{\widehat{\bm \Sigma}}
\newcommand{\eh}{\bm H'}
\newcommand{\norms}[1]{\left\lVert#1\right\rVert_{\esp}}
\newcommand{\normh}[1]{\left\lVert#1\right\rVert_{\eh}}
\newcommand{\normg}[1]{\lVert#1\rVert_{\tt{M}}}

\usepackage{xcolor}

\newtheorem{theorem}{Theorem}
\newtheorem{lemma}[theorem]{Lemma}
\newtheorem{corollary}[theorem]{Corollary}
\newtheorem{proposition}[theorem]{Proposition}
\theoremstyle{remark}
\newtheorem{remark}[theorem]{Remark}

\newcommand{\cp}{c_{\operatorname{p}}}
\newcommand{\cf}{c_{\operatorname{f}}}

\AddToHook{env/lemma/begin}{\crefalias{theorem}{lemma}}
\AddToHook{env/corollary/begin}{\crefalias{theorem}{corollary}}
\AddToHook{env/proposition/begin}{\crefalias{theorem}{proposition}}
\AddToHook{env/remark/begin}{\crefalias{theorem}{remark}}

\title{Compute-Optimal Pretrain–Fine-tune
in Ridge Gradient Descent}

\author{
Alex Buna \thanks{Department of Statistics, University of Oxford, UK} \and
Fanghui Liu \thanks{School of Mathematical Sciences, Institute of Natural Sciences and MOE-LSC, Shanghai Jiao Tong University, China. } \and
Patrick Rebeschini \footnotemark[1]
}
\date{}

\begin{document}

\maketitle
\renewcommand{\thefootnote}{}\footnotetext{Email addresses of corresponding authors:
Alex Buna (\texttt{alex.bunamarginean@spc.ox.ac.uk}),
Fanghui Liu (\texttt{fanghui.liu@sjtu.edu.cn}),
Patrick Rebeschini (\texttt{patrick.rebeschini@stats.ox.ac.uk}).}
\renewcommand{\thefootnote}{\arabic{footnote}}

\begin{abstract}
  Pretraining followed by fine-tuning introduces a compute-allocation problem: under a fixed training budget, compute spent improving the upstream objective reduces the compute available for downstream adaptation. Despite its practical importance, the theoretical understanding of this trade-off remains limited, even in simple models. In this paper, we cast this allocation as a compute-split problem under a two-stage pretrain–fine-tune procedure with fixed total optimisation budget, using regularised least squares trained by gradient descent as a tractable setting. We characterise the optimal split under data-dependent evaluation geometries induced by the fine-tuning problem. Our results show that the allocation depends on how pretraining directions affect fine-tuning predictions and how fine-tuning shifts are seen through downstream data geometry. In particular, the relevant quantities are determined by prediction-relevant spectral components of the pretraining and fine-tuning empirical covariances. Technically, the analysis relies on a basis-invariant, eigenspace-level spectral decomposition, together with perturbative control of the non-commuting pretraining and fine-tuning dynamics.
\end{abstract}

\section{Introduction}
\label{sec:intro}

A central paradigm in modern machine learning is pretraining followed by fine-tuning: a model is first trained on an upstream dataset and subsequently adapted to a downstream task \citep{kumar2022fine,dayi2025gradient}. This two-phase procedure underlies many successful systems in language \citep{devlin2019bert,thoppilan2022lamda} and vision \citep{kumar2022fine}, and fine-tuning may either update all parameters or use parameter-efficient mechanisms such as adapters and LoRA \citep{houlsby2019parameter,hu2022lora}. Computationally, however, the two stages are coupled: for a fixed training budget, compute spent improving the upstream objective is compute not spent adapting to the downstream~task.

This trade-off is procedural rather than purely statistical. Existing theory of pretraining and fine-tuning largely asks when upstream data helps downstream prediction, how task similarity affects transfer, or how pretraining changes generalisation \citep{shachaf2021theoretical,ge2023provable,wu2022power}. We instead fix the two empirical objectives and ask how a finite optimisation budget should be allocated between them. %
A brute-force approach would save multiple pretraining checkpoints, fine-tune each one, and select the best downstream result; a similar intermediate-checkpoint fine-tuning approach is used in recent empirical studies to probe the relation between pretraining and fine-tuning \citep{sun2025amuro}. Such a sweep is informative but expensive, since every candidate checkpoint requires a separate downstream training run. Our objective is to examine the allocation problem within a paradigmatic, analytically tractable framework, and to identify which upstream--downstream geometric quantities determine the optimal split.

To this end, we study this trade-off in ridge-regularised linear regression: suppose the upstream and downstream data are $(\bm X,\bm y) \in\mathbb{R}^{n\times d}\times \mathbb{R}^n$ and $(\bm X',\bm y') \in\mathbb{R}^{n'\times d}\times \mathbb{R}^{n'}$, respectively. For any ridge parameters $\gamma,\gamma'>0$, we consider the pretraining $R_{\operatorname{pre}}(\bm w)=\frac{1}{2n}\|\bm X\bm w-\bm y\|_2^2+\frac{\gamma}{2}\|\bm w\|_2^2$ and fine-tuning $R_{\operatorname{fine}}(\bm w)=\frac{1}{2n'}\|\bm X'\bm w-\bm y'\|_2^2+\frac{\gamma'}{2}\|\bm w\|_2^2$ empirical objectives, with corresponding minimisers $\ww$ and $\ww'$, respectively. For fixed per-iteration costs $\cp>0$ and $\cf>0$ of pretraining and fine-tuning, respectively, running gradient descent (GD) on $R_{\operatorname{pre}}$ for $k$ iterations and then on $R_{\operatorname{fine}}$ for an additional $\ell$ iterations incurs a computational cost of $\cp k+\cf\ell$, which we constrain to equal a fixed total compute budget $B$. This gives the computational question studied in this paper:
\begin{center}
    \emph{Given fixed total compute budget $B$ and per-iteration costs $\cp\ ,\cf$\ ,\\ when should we switch from pretraining to fine-tuning during optimisation?}
\end{center}
To avoid non-integrality issues, we assume that, after choosing a sufficiently fine compute unit, $\cp$ and $\cf$ are positive integers. For simplicity, we focus on budgets $B$ divisible by both $\cp$ and $\cf$, so that $k=0$ and $\ell=0$ are feasible allocations. Thus, there exists a positive integer $T$ such that
\begin{align*}
    B=TL,\quad\text{where } L:=\operatorname{lcm}(\cp,\cf),
\end{align*}
where $\operatorname{lcm}(\cdot,\cdot)$ denotes the least common multiple. Then, every exact-budget schedule $(k,\ell)$ with $\cp k+\cf \ell=B$ is uniquely represented by a feasible pretraining-compute allocation
\begin{align}
\label{eq:representation}
    t \in [T]:=\{0,1,2,\ldots,T\},\quad\text{with}\quad (k(t),\ell(t))=\left(\frac{L}{\cp}t,\frac{L}{\cf}(T-t)\right) .
\end{align}
Many practical adaptation methods are explicitly represented as a task-specific update on top of a shared base model: adapters, LoRA, delta tuning, and task vectors all adopt this viewpoint \citep{houlsby2019parameter,hu2022lora,ding2022delta,ilharco2022editing}. This motivates using %
\begin{align*}
    \bm \Delta:=\ww'-\ww
\end{align*}
as the schedule-independent transfer update from the pretraining optimum to the fine-tuning optimum, and as a natural objective for revealing the trade-off we study in this paper.

A split $(k(t),\ell(t))$ with $t\in[T]$ does not produce $\bm\Delta$ exactly. It first reaches after $k$ iterations the pretraining checkpoint $\bm w_{k(t)}$ and then fine-tunes from that checkpoint for $\ell$ iterations, reaching $\bm w'_{\ell(t)}$. The update induced by this finite-budget two-stage algorithm is therefore
\begin{align*}
    \bm\Delta_{t}:=\bm w'_{\ell(t)}-\bm w_{k(t)}.
\end{align*}
We aim to identify the allocation $t$ that allows $\bm \Delta_t$ to best approximate the benchmark update $\bm \Delta$.

\textbf{Necessity of a fixed anchor:} %
We use $\ww$ as an anchor in order to compare all allocations against the same transfer update $\bm\Delta$. If the checkpoint $\bm w_{k(t)}$ were used as the anchor instead, the ideal update would be $\ww'-\bm w_{k(t)}$, and the mismatch reduces to
$\bm\Delta_t-(\ww'-\bm w_{k(t)})=\bm w'_{\ell(t)}-\ww'$, measuring only how close the fine-tuning iterate is to the fine-tuning optimum. It would no longer compare the quality of the transfer update across different pretrain--fine-tune splits. Anchoring at $\ww$ thus gives~a~fixed~comparator~that~is~independent~of~the~split.

Once updates are anchored at $\ww$, the natural way to evaluate a split $(k(t),\ell(t))$ is through the empirical fine-tuning risk of the transferred model $\ww+\bm\Delta_t$\ , which can be expressed (see \cref{lemma: ridge connection})~as
\begin{align*}
    R_{\operatorname{fine}}(\ww+\bm \Delta_t)-R_{\operatorname{fine}}(\ww' )
    &=
    \frac{1}{2}\|\bm \Delta_t-\bm \Delta\|_{\esp}^2
    +\frac{\gamma'}{2}\norm{\bm \Delta_t-\bm \Delta}^2
    =
    \frac{1}{2}\|\bm \Delta_t-\bm \Delta\|_{\esp+\gamma'\bm I}^2 \ ,
\end{align*}
where $\esp=\frac{1}{n'}(\bm X')^\top \bm X'\in\mathbb{R}^{d\times d}$ denotes the downstream empirical covariance. This identity motivates evaluating the \emph{update mismatch}  $\|\bm \Delta_t-\bm \Delta\|^2_{\tt{M}}$ under two different geometries $\texttt{M}\in\{\bm H',\esp\}$, where $\eh :=\esp+\gamma'\bm I$. The \emph{norm} $\|\cdot\|_{\eh}$\ , induced by the positive definite fine-tuning Hessian $\eh$, exactly controls the empirical excess risk above. The \emph{seminorm} $\|\cdot\|_{\esp}$ ignores directions in $\ker(\bm X')$, which are unidentifiable from the downstream sample and do not affect its empirical predictions. Consequently, we formalise the allocation problem as the minimisation of
\begin{align*}
    f_{\tt M}:[T]\rightarrow[0,\infty),\quad f_{\tt M}(t)=\|\bm\Delta_t-\bm\Delta\|_{\tt M}^2,
    \quad
    {\tt M}\in\{\bm H',\esp\}.
\end{align*} 

\subsection{Our contributions and technical challenges}
\label{sec:contribution}

\begin{figure}[t]
  \centering
  \includegraphics[width=\textwidth]{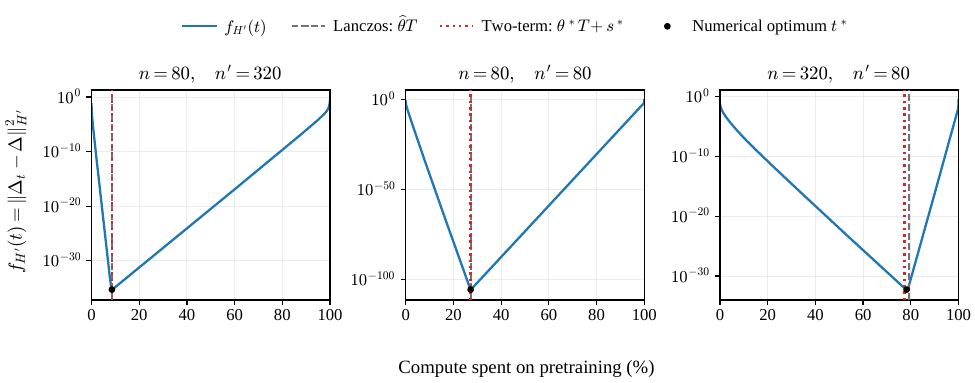}
  \caption{
We generate random-design linear regression examples in dimension $d=400$, with $(n,n')=(80,320),(80,80),(320,80)$ pretraining and fine-tuning samples, respectively (left/middle/right). For each feasible allocation $t\in\{0,\ldots,T\}$, we evaluate the closed-form full-batch GD dynamics for $k(t)$ pretraining iterations followed by $\ell(t)$ fine-tuning iterations, as in representation (\ref{eq:representation}). Let $\bm \Delta_t$ be the algorithm-induced update and $\bm \Delta=\ww'-\ww$ the ideal transfer update. The blue curve shows $f_{\eh}(t)=\|\bm \Delta_t-\bm \Delta\|_{\eh}^2$ against the percentage of compute spent on pretraining, $100t/T$. The panels illustrate how sample sizes and the resulting empirical geometry affect the optimal allocation (c.f.\ \hyperref[remark: sample sizes]{Remark~\ref*{remark: sample sizes}}). The black dot marks the numerical optimum found by evaluating every feasible allocation. The red dotted line marks the predicted $\theta^* T+s^*$ in (\ref{def theta}), computed using exact spectral and solution-level quantities according to \cref{thm:localisationl2}. The grey dashed line marks the estimated leading prediction $\widehat\theta T$, based on Lanczos iterations and described in \cref{subsec:complexity-separation}. This estimate $\widehat{\theta}$ is computed without the ridge solutions or full spectral decompositions. For reproducibility, further experimental details have been included in \hyperref[app:experiments]{Appendix~\ref*{app:experiments}}.
}
  \label{fig:linear}
\end{figure}

\textbf{Localisation of the optimal split.}
Our main results (\cref{thm:localisationl2} and~\ref{thm:localisation}) show that, for the reparametrisation (\ref{eq:representation}), for sufficiently large $T$, any global minimiser $t^*$ of $f_{\tt M}$ satisfies
\begin{align}
        \left|t^*-\left(s^* + \theta^* T\right)\right|\leq 1+ O(e^{-\delta T}),\qquad\text{where}\quad\theta^*:=\frac{\frac1\cf \ln(1-\eta'\lambda'_{j^*})}{\frac1\cp\ln(1-\eta\lambda_{i^*})+\frac1\cf\ln(1-\eta'\lambda'_{j^*})}. \label{def theta}
    \end{align}
The \emph{leading fraction} $\theta^*$ depends on the step-sizes $\eta,\eta'>0$ of GD in each of the two phases, the normalisation with respect to the two costs $\cp,\cf$, and two active bottlenecks $\lambda_{i^*},\lambda'_{j^*}$\ . The pretraining bottleneck $\lambda_{i^*}$ is the smallest eigenvalue among those whose upstream eigenspaces have a non-zero (visible) projection of $\ww$ in $\tt{M}$-geometry, while the fine-tuning bottleneck $\lambda'_{j^*}$ is the smallest eigenvalue among those whose downstream eigenspaces have a non-zero (visible) projection of $\bm\Delta$ in $\tt{M}$-geometry. The offset $s^*$ is the next-order, $T$-independent correction. It depends on the scale and alignment of these smallest upstream and downstream visible projections. Furthermore, in \cref{cor: kl} we obtain the corresponding expressions for the optimal split
\begin{align*}
    (k(t^*),\ell(t^*))\approx \left(\theta^*\frac{B}{\cp},(1-\theta^*)\frac{B}{\cf}\right).
\end{align*}
Thus, the same upstream task can have different optimal allocations for different downstream tasks, and vice versa. This is illustrated in \cref{fig:linear}, where changing only the sample sizes moves the predicted optimal compute allocation from early to late pretraining.

Complementing these results, \cref{thm:improvement} shows that any fixed interior fraction $\widehat\theta\in(0,1)$ gives an update mismatch exponentially smaller in $T$ than allocations with only a constant amount of pretraining or fine-tuning, and the guaranteed improvement is maximised at $\widehat\theta=\theta^*$\ . Importantly, computing the offset $s^*$ is not required for this exponential improvement, and this motivates the estimation approach in \cref{subsec:complexity-separation}, also illustrated in \cref{fig:linear}. While evaluating $\theta^*T+s^*$ requires solution-level information, the leading fraction $\theta^*$ depends only on $\lambda_{i^*}$ and $\lambda'_{j^*}$. We estimate these quantities using matrix-free Lanczos iterations, without computing the ridge solutions or full spectral decompositions. With a fixed number of Lanczos iterations, estimating these bottlenecks costs \(O((n+n')d)\). This provides a potential alternative to a checkpoint sweep over many candidate~splits.

\textbf{An eigenspace-level perturbation analysis for non-commuting objectives.}
Technically, the proof uses spectral projectors associated with distinct eigenvalues, rather than an arbitrary eigenvector basis. This makes the active sets basis-invariant when eigenvalues have multiplicity and allows us to control the non-commuting cross-terms between the pretraining and fine-tuning ridge Hessians. Under the downstream prediction seminorm $\|\cdot\|_{\widehat{\bm\Sigma}'}$\ , the result requires a `no-kernel-hitting' condition because $\widehat{\bm\Sigma}'$ can vanish on non-predictive directions. \cref{prop:counterexample_nkh} shows why such a condition is necessary, while \cref{prop:random design,prop:commutative} give common settings where it holds.

\subsection{Related work}
\textbf{Statistical theory of pretraining and fine-tuning.}
Most theoretical work on pretraining and fine-tuning studies statistical transfer: when upstream data improves downstream prediction, how task similarity affects sample complexity, or how covariate shift changes the value of pretraining \citep{shachaf2021theoretical,ge2023provable,wu2022power}. The closest works to ours are linear-model analyses. In linear-teacher settings, \citet{shachaf2021theoretical} show that the relevant similarity notion depends on the source task, target task, and covariance structure of the target data. For linear regression under covariate shift, \citet{wu2022power} derive instance-dependent guarantees for pretraining followed by fine-tuning with online SGD. These works identify when upstream data can statistically help downstream prediction. Our work is complementary: we fix the two empirical objectives and ask where a finite optimisation budget should switch from one objective to the other.

\textbf{Compute allocation.}
Our question is also connected to compute-optimal training. Scaling-law works study how finite compute should be allocated across axes such as model size, data, and optimisation time \citep{brown2020language,hoffmann2022training,paquette20244+}. Transfer settings introduce another allocation axis: upstream training versus downstream adaptation. For instance, for protein language models, \citet{cheng2024training} find a non-trivial switching trade-off, allocating approximately 10–20\% of the fixed compute budget to causal language modelling before switching to masked language modelling. Other empirical transfer-scaling studies show that downstream behaviour depends on the amount and alignment of pretraining and fine-tuning data \citep{hernandez2021scaling,isik2024scaling}, and LLM fine-tuning studies find that performance depends on model size, pretraining and fine-tuning data, and the fine-tuning method \citep{zhang2024scaling}. Such observations motivate a theory in which the optimal split depends on both upstream and downstream geometry, rather than on upstream loss alone.

\textbf{Checkpoint selection and update-based adaptation.}
A practical way to study the pretrain--fine-tune interface is to fine-tune several intermediate pretraining checkpoints.  The work of \cite{sun2025amuro} takes this approach for LLMs and shows that downstream behaviour after supervised fine-tuning can reveal effects of pretraining that are not apparent from the pretraining checkpoint alone. Our ridge model abstracts this situation into a discrete compute-allocation problem. The update-based formulation is also motivated by parameter-efficient and arithmetic views of adaptation: adapters and LoRA learn small task-specific modifications \citep{houlsby2019parameter,hu2022lora}, delta tuning frames adaptation as a parameter delta \citep{ding2022delta}, and task arithmetic manipulates differences between base and fine-tuned weights \citep{ilharco2022editing}. This motivates evaluating the quality of the finite-budget update $\bm\Delta_t$, not only the final fine-tuned iterate.

\textbf{Ridge dynamics.}
Ridge-regularised least squares yields closed-form GD trajectories through the spectrum of the empirical covariance \citep{hastie2009elements,bach2024learning}. Spectral analyses have been used to characterise mode-wise learning dynamics in linear and deep-linear models \citep{saxe2013exact,tarmoun2021understanding}. Related work connects optimisation paths to explicit $\ell_2$ regularisation and spectral filtering \citep{rosasco2005spectral,gerfo2008spectral,suggala2018connecting,zhu2021algorithmic,ali2019continuous,wu2025risk}. %
Recent advances also show that fixed-feature models, similar to the setting under consideration, can remain informative beyond that formal setting via spectral predictions. For instance, \citet{li2026functional} derive a functional scaling law under power-law spectral assumptions and verify empirically that it predicts LLM loss trajectories under unseen schedules; and \citet{karkada2026predicting} empirically show that covariance-derived kernel eigenvalues predict the order in which feature-learning MLPs learn Hermite-polynomial~targets. %

\subsection{Notation and organisation of the paper}
For an integer $m\geq 0$, let $[m]:=\{0,1,\ldots,m\}$, with the convention that $[-1]=\emptyset$. We denote the zero vector by $\bm{0}$ and the identity matrix by $\bm I$. For vectors $\bm{u},\bm{v}\in\mathbb{R}^d$ and a symmetric positive semidefinite matrix $\bm{A}\in\mathbb{R}^{d\times d}$, we use the notations $\langle \bm{u},\bm{v}\rangle_{\bm{A}}=\bm{u}^\top \bm{A}\bm{v}$ and $\smallnorm{\bm{u}}_{A}=\langle \bm{u},\bm{u}\rangle_{\bm{A}}^{1/2}$. The Euclidean and operator norms will be denoted by $\smallnorm{\bm{v}}_2$ and $\smallnorm{\bm{A}}_2$, respectively. 

\cref{sec:preliminaries} formally introduces the model, algorithm, and evaluation metric. \cref{sec:localisation} presents our main theoretical findings on the localisation of the optimal allocation, with a  proof sketch of one of these results in \hyperref[proofsketch]{Appendix~\ref*{proofsketch}}. All missing proofs, deferred to \hyperref[app:localisation]{Appendix~\ref*{app:localisation}}, rely on additional notation in  \hyperref[additionalnotation]{Appendix~\ref*{additionalnotation}}. Experimental details for \cref{fig:linear} can be found in \hyperref[app:experiments]{Appendix~\ref*{app:experiments}}.

\section{Problem settings}
\label{sec:preliminaries}

This section presents our framework: ridge regression optimised by GD, and~the~evaluation~metrics.

\textbf{Ridge regression and gradient descent.}
Let $(\bm X, \bm y)\in\mathbb{R}^{n\times d}\times \mathbb{R}^n$ and $(\bm X', \bm y')\in\mathbb{R}^{n'\times d}\times \mathbb{R}^{n'}$  denote the  pretraining and  fine-tuning datasets, respectively. For fixed ridge parameters $\gamma, \gamma'>0$, define the ridge-regularised empirical risks
\begin{align*}
    R_{\operatorname{pre}}(\bm w)=\frac{1}{2n}\norm{\bm X \bm w- \bm y}^2+\frac{\gamma}{2}\norm{\bm w}^2,\quad R_{\operatorname{fine}}(\bm w)=\frac{1}{2n'}\norm{\bm X' \bm w-\bm y'}^2+\frac{\gamma'}{2}\norm{\bm w}^2.
\end{align*}
Consider the associated empirical covariances and positive definite ridge Hessians
\begin{align*}
    \widehat{\bm \Sigma}=\frac{1}{n}\bm X^\top \bm X,\quad \bm H=\widehat{\bm \Sigma}+\gamma \bm I,\quad
    \widehat{\bm \Sigma}'=\frac{1}{n'} \bm X'^\top \bm X',\quad \bm H'=\widehat{\bm \Sigma}'+\gamma' \bm I\,.
\end{align*}
The pretraining and fine-tuning unique global minimisers~are~given~by
\begin{align*}
    \bm \ww:=\arg\min_{\bm w} R_{\operatorname{pre}}(\bm w)=\frac{1}{n} \bm H^{-1}\bm X^\top \bm y,\quad \bm \ww':=\arg\min_{\bm w} R_{\operatorname{fine}}(\bm w)=\frac{1}{n'}(\bm H')^{-1}(\bm X')^\top \bm y'.
\end{align*}

Recall that we aim to recover the displacement $\bm\Delta=\ww'-\ww$. For this, consider a fixed computational budget $B>0$ and step-sizes $0<\eta<\norm{\bm H}^{-1}$, $0<\eta'<\norm{\bm H'}^{-1}$. For any split $(k,\ell)$ with $\cp k+\cf\ell=B$, define the following two-phase GD procedure:%
\begin{enumerate}
    \item \textbf{Pretraining phase:} Initialise $\bm w_0=\bm{0}$ and run
     $   \bm w_{s+1}=\bm w_s-\eta\nabla R_{\operatorname{pre}}(\bm w_s)$ for $s\in[k-1]$.
    \item \textbf{Fine-tuning phase:} Initialise $\bm w_0'= \bm w_{k}$ and run 
     $ \bm w'_{s+1}=\bm w'_s-\eta'\nabla R_{\operatorname{fine}}(\bm w'_s)$ for $s\in[\ell-1]$.
\end{enumerate}
Accordingly, under the representation (\ref{eq:representation}), a natural estimator for $\bm{\Delta}$ is the update $\bm\Delta_t\!=\!\bm w'_{\ell(t)}\!-\bm w_{k(t)}$.

\textbf{Evaluation metrics.}
Recalling the discussion from Section \ref{sec:intro} on the necessity of a canonical anchor, once one evaluates such an update anchored at $\ww$, the natural performance criterion is the empirical fine-tuning risk of the transferred model $\ww+\bm\Delta_t$. In ridge linear regression, this connection is exact:

\begin{lemma}
    \label{lemma: ridge connection}
    The following fine-tuning empirical excess risk identity holds:
    \begin{align*}
        R_{\operatorname{fine}}(\ww+\bm \Delta_t)-R_{\operatorname{fine}}(\ww' )=\frac{1}{2}\norms{\bm \Delta_t-\bm \Delta}^2+\frac{\gamma'}{2}\norm{\bm \Delta_t-\bm \Delta}^2=\frac{1}{2}\normh{\bm \Delta_t-\bm\Delta}^2.
    \end{align*}
\end{lemma}

One goal of our work is to understand how one can choose $t$ so that the anchored update fine-tuning risk  $R_{\operatorname{fine}}(\ww+\bm{\Delta}_t)$ is as close as possible to $R_{\operatorname{fine}}(\ww')$. With the viewpoint of anchored update given by Lemma \ref{lemma: ridge connection}, we measure the quality of this approximation via a parameter-space distance
\begin{align*}
    f_{\bm H'}(t):=\normh{\bm\Delta_t-\bm\Delta}^2\quad\text{and}\quad f_{\esp}(t):=\norms{\bm{\Delta}_t- \bm{\Delta}}^2 = \frac{1}{n'}\norm{\bm X'(\bm{\Delta}_t- \bm{\Delta})}^2.
\end{align*}

In general, for ${\tt{M}}\in\{\eh,\esp\}$ we write $f_{\tt M}(t)=\smallnorm{\bm \Delta_t-\bm \Delta}_{\tt M}^2$. %
The set of minimisers 
\begin{align*}
    \mathcal T^* := \underset{t\in[T]}{\arg\min} f_{\tt M}(t)
\end{align*}
represents the optimal allocation of compute between pretraining and fine-tuning that best approximates the ideal transfer shift under the fine-tuning geometry or the $\bm H'$-induced geometry. %

\section{Localisation of the optimal pretrain/fine-tune split}
\label{sec:localisation}

Before delivering our results, we give an alternative formulation for $f_{\tt M}$ involving $\ww$, $\bm \Delta$, $\bm H$, $\bm H'$\ :
\begin{lemma}
\label{lemma f(t) form}
    For every split $(k(t),\ell(t))$, the pretraining and fine-tuning GD iterates satisfy
    \begin{align*}
       \bm w_{k(t)}=(\bm I-(\bm I-\eta \bm H)^{k(t)}) \bm \ww,\quad \bm w'_{\ell(t)}= \bm \ww' + (\bm I-\eta'\bm H')^{\ell(t)}(\bm w_{k(t)}- \bm \ww')\,.
    \end{align*}
    Consequently, under $\smallnorm{\cdot}_{{\tt M}}$ (using either $\| \!\cdot\! \|_{\bm H'}$ or $\norms{\cdot}$), we have
    \begin{align*}
        f_{\tt M}(t)=\normg{(\bm I-(\bm I-\eta'\bm H')^{\ell(t)})(\bm I-\eta \bm H)^{k(t)} \bm \ww - (\bm I-\eta'\bm H')^{\ell(t)}\bm{\Delta}}^2.
    \end{align*}
\end{lemma}
Note that, due to the step-sizes choice, the matrices $\bm I-\eta \bm H$ and $\bm I-\eta'\bm H'$ are positive definite.

\begin{remark} 
Let us consider the extreme cases: $t=0$ represents no pretraining, so $\bm{\Delta}_0= \bm w'_{\ell(0)}$ and $f_{\tt M}(0)=\normg{\bm \ww-(\bm I-\eta'\bm H')^{B/\cf} \bm \ww'}^2$\ ; $t=T$ corresponds to no fine-tuning, so $\bm{\Delta}_T=0$ and $f_{\tt M}(T)=\normg{\bm{\Delta}}^2$.
While these two quantities are not immediately comparable, we prove in \cref{thm:improvement} that, at the global minima identified in \cref{thm:localisationl2}, the value of $f_{\eh}$ represents an exponential improvement over very little pretraining (small $k$) or fine-tuning (small $\ell$).

\end{remark}

\subsection{Warm-up with special cases of $t^*=0$ or $t^*=T$}
\label{sec:spt}
As a starting point, in this section we discuss three degenerate cases that arise due to the interplay between $\ww$, $\bm \Delta$ and the spectral structure of the positive definite Hessians $\bm H$ and $\bm H'$.

Let $0<\lambda_1<\lambda_2<\cdots<\lambda_r$ be the \emph{distinct} eigenvalues of $\bm H$, with corresponding orthogonal eigenspace projectors $\{\bm P_i\}_{i=1}^r$. Thus, if $\bm U_i$ contains as columns an orthonormal basis for the eigenspace associated with $\lambda_i$, then $\bm P_i=\bm U_i\bm U_i^\top$. Define $\{\lambda'_j,\bm P'_j\}_{j=1}^{r'}$ analogously for $\bm H'$, so that
\begin{align*}
    \bm H=\sum_{i=1}^r \lambda_i \bm P_i, \quad
    \bm H'=\sum_{j=1}^{r'}\lambda'_j\bm P'_j,\quad
    \sum_{i=1}^r\bm P_i=\sum_{j=1}^{r'}\bm P'_j=\bm I\,.
\end{align*}

Consider the \emph{active} sets for pretraining and fine-tuning through the geometry induced by $\smallnorm{\cdot}_{{\tt M}}$:
\begin{align*}
    \mathcal{I}:=\{i:\normg{\bm P_i\ww}>0\},\quad
    \mathcal{I}':=\{j:\normg{\bm P'_j\bm{\Delta}}>0\}\,.
\end{align*}
For instance, under the downstream prediction seminorm $\norms{\cdot}$\ , $\mathcal{I}$ contains the pretraining eigenspaces whose residual signal is visible to fine-tuning prediction, while $\mathcal{I}'$ contains the fine-tuning shift eigenspaces that are visible in downstream data geometry.

We deliberately work with projectors onto eigenspaces of distinct eigenvalues, rather than with an arbitrary eigenvector basis. This distinction matters when eigenvalues have multiplicity: individual eigenvectors are not canonical, but the projector $\bm P_i$ onto the full eigenspace is basis-invariant. The active sets are therefore defined through eigenspace-level quantities.

We now outline three degenerate cases, when either one or both of these active sets are empty.

\begin{theorem}[Special cases for $t^*$] 
\label{thm:pathologic}
Let $t^*$ be a global minimiser of $f_{\tt M}:[T]\rightarrow[0,\infty)$ defined as $f_{\tt M}(t):=\normg{\bm{\Delta}_t-\bm{\Delta}}^2$\ .
    \begin{enumerate*}[label=(\roman*),itemjoin={\quad}]
        \item If $\mathcal{I}\neq\emptyset$, $\mathcal{I}'=\emptyset$, then $T\in \mathcal T^*$.
        \item If $\mathcal{I}=\emptyset$, $\mathcal{I}'\neq\emptyset$, then $0\in \mathcal T^*$.
        \item If $\mathcal I=\mathcal{I}'=\emptyset$, then $\mathcal T^*=[T]$.
    \end{enumerate*}
\end{theorem}

\begin{remark}
Case (i) indicates that the pretraining solution $\ww$ has components that matter for the fine-tuning task, but the true task shift $\bm{\Delta}$ is invisible under $\normg{\cdot}$. In other words, the optimal allocation is full pretraining and no fine-tuning.
Case (ii) is the opposite of Case (i): the pretraining solution $\ww$ is completely irrelevant for the fine-tuning task, but the fine-tuning shift does matter. So, we take a no pretraining but full fine-tuning strategy. Case (iii) is the trivial case.
\end{remark}

\cref{thm:pathologic} characterises degenerate regimes where either pretraining or fine-tuning is completely useless, and a boundary allocation $t=0$ or $t=T$ is optimal. From now on, we will assume  $\mathcal{I}, \mathcal{I}' \neq \emptyset$, in which case $t^*$ can be an interior point, as we show next.

\subsection{General cases: $\mathcal I,\mathcal I'\neq\emptyset$ and fine-tuning ridge Hessian evaluation}
\label{sec:genl2}
The results for this general case are more complex.
For ease of description, we first state our results for $\texttt M=\eh$, which are clearer, and then provide in \cref{sec:genf} the corresponding results for $\texttt M=\esp$.

\begin{theorem}[General case under the $\bm H'$ norm]
\label{thm:localisationl2}
    Assume $\mathcal{I}, \mathcal{I}' \neq \emptyset$ under the $\bm H'$ norm and consider $i^*=\min \mathcal{I}$, $j^*=\min \mathcal{I}'$, and $\theta^*$ as defined in (\ref{def theta}).
    Then, there exist an offset $s^*\in\mathbb{R}$, a rate $\delta>0$, and $C,T_0>0$, all functions of the problem parameters $L,\cp,\cf,\gamma,\gamma',\es,\esp,\ww,\bm{\Delta},\eta,\eta'$,  such that the following holds: for all $T\geq T_0$,  the evaluation metric $f_{\bm H'}:[T]\rightarrow[0,\infty)$, $f_{\bm H'}(t)=\|\bm\Delta_t-\bm\Delta\|^2_{\bm H'}$ has at most two global minimisers, each $t^*\in\mathcal T^*$ satisfying
    \begin{align*}
        \left|t^*-(s^* + \theta^*T)\right|\leq 1+ Ce^{-\delta T}.
    \end{align*}
\end{theorem}
As the proof of \cref{thm:localisationl2} reveals, for sufficiently large $T$, the relaxation of $f_{\bm H'}$ to $[0,T]$ has a unique global minimiser, and every discrete global minimiser is its floor or ceiling. Thus, up to one feasible allocation step, $t^*$ is approximated by two terms with an exponentially decaying error. The term linear in $T$ is determined by the \emph{contraction rates} $-\ln(1-\eta\lambda_{i^*})$ and $-\ln(1-\eta'\lambda_{j^*}')$ of smallest active eigenvalues $\lambda_{i^*}$ and $\lambda'_{j^*}$ of $\bm H$ and $\bm H'$, respectively, and the compute normalisation $1/\cp$ and $1/\cf$. The offset $s^*$ is independent of $T$, hence, as $T\rightarrow\infty$, $t^*/T$ converges to the \emph{leading fraction} $\theta^*$. The corresponding results for the optimal iteration counts $k(t^*)$ and $\ell(t^*)$, accompanied by a geometric interpretation, can be found in \cref{optimal iteration counts}.

The offset $s^*$ is the correction to the leading allocation $\theta^*T$, and its explicit expression can be found in \hyperref[additionalnotation]{Appendix~\ref*{additionalnotation}}. Geometrically, it is quantified by (i) the misalignment in $\eh$-geometry between the two slow `bottleneck' directions via $\langle \bm P_{i^*}\ww, \bm P'_{j^*}\bm{\Delta}\rangle_{\eh}$\ , induced by the non-commutativity of $\es$ and $\esp$; (ii) the strength of the dominant pretraining $\smallnorm{\bm P_{i^*}\ww}_{\eh}^2$ and fine-tuning $\smallnorm{\bm P'_{j^*}\bm{\Delta}}_{\eh}^2$ components.

The rate $\delta$, whose explicit expression can also be found in \hyperref[additionalnotation]{Appendix~\ref*{additionalnotation}}, depends on the separation between the bottleneck contraction rates and the remaining active rates. Small spectral separations can therefore delay the budget regime in which the two-term approximation becomes accurate.

    Next, \cref{thm:improvement} shows that the value of $f_{\eh}$ at interior points represents an exponential improvement (in $T\propto B$) over extreme allocations with a fixed number $M$ of units assigned to either phase.

\begin{theorem}[Exponential improvement over extreme allocations]
\label{thm:improvement}
    Recall the setting of \cref{thm:localisationl2}. Fix any $\widehat{\theta}\in(0,1)$ and define the feasible allocation $\widehat{t}\in[T]$ and rate $m(\widehat{\theta})>0$ as
    \begin{align*}
        \widehat t \in \arg\min_{t\in [T]}\left|t-\widehat{\theta}T\right|, \qquad m(\widehat \theta):=\min\left\{-\frac{L\widehat{\theta}}{\cp}\ln(1-\eta\lambda_{i^*}),-\frac{L(1-\widehat{\theta})}{\cf}\ln(1-\eta'\lambda'_{j^*})\right\}.
    \end{align*}
    Fix any integer $M\geq 1$. Then, there exists $C_1>0$ and $T_0>2M$, depending on $M$ and the fixed problem parameters, but independent of $\widehat{\theta}$, such that for any $T\geq T_0$,
    \begin{align*}
        f_{\eh}(\ \widehat{t}\ )\leq C_1 \exp(-2m(\widehat{\theta})T)\cdot \min\{f_{\eh}(t): t\in \{0,\ldots,M\}\cup \{T-M,\ldots,T\}\}.
    \end{align*}
    The decay rate $m(\widehat{\theta})$ is maximised for $\widehat{\theta}=\theta^*$ defined in (\ref{def theta}).
\end{theorem}

\begin{remark}
    Although the offset $s^*$ in \cref{thm:localisationl2} shifts the optimal $t^*$ away from $\theta^*T$, this offset depends on solution-level information, including $\ww$ and $\ww'$. Thus, evaluating $t^*$ can be as computationally intensive as solving each ridge problem separately.
    \cref{thm:improvement} guarantees an exponential improvement over extreme allocations \emph{without computing the offset $s^*$}.  Since an estimate $\widehat\theta\approx\theta^*$ retains the improvement with $m(\widehat{\theta})$ close to $m(\theta^*)$, one can instead use the split induced by an integer nearest to $\widehat\theta T$. \cref{subsec:complexity-separation} describes how to perform such an estimation of $\theta^*$ in $O((n+n')d)$ time using matrix-free Lanczos iterations.

\end{remark}

\subsubsection{Optimal iteration counts}
\label{optimal iteration counts}

We now return to the original iteration counts $k$ and $\ell$. Using the representation (\ref{eq:representation}), \cref{thm:localisationl2} gives the following corollary, in which the $O(1)$ terms absorb the offset and integral allocation error:

\begin{corollary}
    \label{cor: kl} Under the assumptions of \cref{thm:localisationl2}, every optimal schedule $(k(t^*),\ell(t^*))$ satisfies
    \begin{align*}
        k(t^*)=\theta^*\frac B\cp+O(1),\quad
        \ell(t^*)=(1-\theta^*)\frac B\cf+O(1).
    \end{align*}
    In particular, as $B=LT\rightarrow\infty$,
    \begin{align*}
        \frac{k(t^*)}{\ell(t^*)}\rightarrow\frac{\ln(1-\eta'\lambda'_{j^*})}{\ln(1-\eta\lambda_{i^*})}.
    \end{align*}
\end{corollary}

\begin{remark}
    Our two-phase algorithm optimises a strongly convex quadratic in each phase. GD progresses faster in high-curvature directions, and slower in low-curvature directions. Thus, the relevant bottlenecks are the flattest active directions, indexed by $i^*$ and $j^*$. Each GD iteration contracts the hardest active component by $1-\eta\lambda_{i^*}$ or $1-\eta'\lambda'_{j^*}$\ , respectively. The inverse logarithmic contraction rates therefore measure the iteration scales for constant progress:
\begin{align*}
    k(t^*)\propto\frac{1}{-\ln(1-\eta\lambda_{i^*})} \quad\text{and}\qquad \ell(t^*)\propto\frac{1}{-\ln(1-\eta'\lambda'_{j^*})},
\end{align*}
with a common proportionality factor fixed by the compute constraint $\cp k(t^*)+\cf\ell(t^*)=B$. Up to the offset and integer allocation, this insight recovers the iteration counts of \cref{cor: kl}.
\end{remark}

\begin{remark}[Sample sizes and compute allocation] 
\label{remark: sample sizes}
For dense full-batch GD, the iteration costs scale as $\cp=\Theta(nd)$ and $\cf=\Theta(n'd)$, so that $\cp/\cf\approx n/n'$. According to \cref{cor: kl}, at leading order, the optimal pretraining compute is:
    \begin{align} 
         \cp k(t^*) \approx \frac{n}{ n+n' \frac{\ln(1-\eta\lambda_{i^*})} {\ln(1-\eta'\lambda'_{j^*})} }\,B.  \label{eq:cpk}
    \end{align}
Thus, holding the contraction rates fixed, increasing $n/n'$ shifts the leading allocation towards pretraining, because each upstream pass becomes relatively more expensive. The ratio of the contraction rates adjusts this sample-size weighting for the relative difficulty of the two phases. In \cref{fig:linear}, the ratio of contraction rates is approximately $2.626/ 2.626/1.003$ in the left/middle/right panels, respectively. In particular, when $n=n'=80$, a fraction of $1/(1+2.626)\approx27.58\%$ of compute is allocated by (\ref{eq:cpk}) to pretraining, close to the numerical optimum of $27.21\%$.
\end{remark}

\subsection{General cases: $\mathcal I,\mathcal I'\neq\emptyset$ under the fine-tuning data geometry}
\label{sec:genf}

We now provide our results on the optimal set $\mathcal T^*$ under the $\esp$-seminorm. Compared to the $\bm H'$ norm evaluation metric case, these results require an additional assumption that we discuss in detail.

\subsubsection{Localisation of $t^*$}
\begin{theorem}[General case under the $\norms{\cdot}$ seminorm]
\label{thm:localisation}
    Assume $\mathcal{I}, \mathcal{I}' \neq \emptyset$ under the $\norms{\cdot}$ seminorm and consider $i^*=\min \mathcal{I}$ and $j^*=\min \mathcal{I}'$ and $\theta^*$ as defined in (\ref{def theta}). Then, there exist an offset $s^*\in\mathbb{R}$, a rate $\delta>0$, an integer burn-in $K\geq 0$, and $C>0$, $T_0\geq K$, all functions of the problem parameters $L$, $\cp$, $\cf$, $\gamma$, $\gamma'$, $\es$, $\esp$, $\ww$, $\bm{\Delta}$, $\eta$, $\eta'$,  such that the following holds: Assume the \emph{no-kernel-hitting}~condition:
    \begin{align}
    \label{nkh}
        \text{for all } t\in[K]: \quad \norms{\ww-\bm w_{k(t)}}\neq 0. \tag{NKH}
    \end{align}
    Then, for all $T\geq T_0$, the evaluation metric $f_{\esp}:[T]\rightarrow[0,\infty)$, $f_{\esp}(t)=\|\bm\Delta_t-\bm\Delta\|^2_{\esp}$ has at most two global minimisers, each $t^*\in\mathcal T^*$ satisfying
    \begin{align*}
        \left|t^*-\left(s^* + \theta^*T\right)\right|\leq 1+ Ce^{-\delta T}.
    \end{align*}
\end{theorem}

These results under the $\smallnorm{\cdot}_{\esp}$ seminorm are very similar to the ones of \cref{thm:localisationl2} under the $\bm H'$ norm, and the expression for $K$, $s^*$, and $\delta$ can be explicitly found in \hyperref[additionalnotation]{Appendix~\ref*{additionalnotation}}. 

Different from the $\|\cdot\|_{\eh}$ evaluation metric, $\esp$ could in principle have $0$ eigenvalues, causing $\smallnorm{\cdot}_{\esp}$ to become a seminorm that vanishes on $\operatorname{ker}(\esp)$. Hence, we make the `no-kernel-hitting' assumption (\ref{nkh}): for a burn-in period $K$, the pretraining GD trajectory $t\mapsto \ww-\bm w_{k(t)}$ does not enter $\operatorname{ker}(\esp)$, remaining visible to downstream prediction. 
As we show next, there are cases where (\ref{nkh}) does not hold and, unlike in the conclusion of \cref{thm:localisation}, $t^*$ may not scale linearly with $T$.

\begin{proposition}
    \label{prop:counterexample_nkh}
    There exist $(\bm X,\bm y),(\bm X', \bm y')\in\mathbb{R}^{2\times 2}\times \mathbb{R}^2$ and $\cp,\cf,\eta,\eta',\gamma,\gamma'>0$, such that, if $T\geq 3$, then the (\ref{nkh}) condition does not hold for any $K\geq 3$, i.e.\, $\|\ww-\bm w_{k(t)}\|_{\esp}=0$ for some $t\in[K]$.
    Additionally, $f_{\esp}(t)$ has a unique global minimiser at $t^*=1$, so that ${t^*}/{T}\rightarrow 0$~as~$T\rightarrow\infty$.
\end{proposition}

However, under a standard random-design sampling mechanism, the condition (\ref{nkh}) holds almost surely, and it is weaker than a commutativity assumption on $\es$ and $\esp$, as we show next.

\begin{proposition}
\label{prop:random design}
Assume that $n'\geq 1$ and the rows of $\bm{X}'\in\mathbb{R}^{n'\times d}$ are drawn from a distribution supported on $\mathbb{R}^d$ that has a density with respect to the Lebesgue measure, and are independent of the (potentially deterministic) pretraining data. Moreover, assume $\ww\neq 0$ (otherwise $\mathcal{I}=\emptyset$, see \cref{thm:pathologic}). Then, almost surely, $\|\ww-\bm w_{k(t)}\|_{\esp}>0$ for every integer $t\geq 0$.
    
\end{proposition}

\begin{proposition}
    \label{prop:commutative}
    Assume that $\es$ and $\esp$ commute. Then (deterministically):
    \begin{enumerate}[label=(\roman*)]
        \item If $\ww\in\operatorname{ker}(\esp)$, then $\smallnorm{\ww-\bm w_{k(t)}}_{\esp}=0$ for all integers $t\geq 0$, but $\mathcal{I}=\emptyset$ (c.f. \cref{thm:pathologic}).
        \item If $\ww\notin\operatorname{ker}(\esp)$, then (\ref{nkh}) holds.
    \end{enumerate}
\end{proposition}

\subsection{Estimating the leading fraction $\theta^*$}
\label{subsec:complexity-separation}

Evaluating the offset $s^*$ in \cref{thm:localisationl2,thm:localisation}, whose explicit expression can be found in \hyperref[additionalnotation]{Appendix~\ref*{additionalnotation}}, requires information about projections of $\ww$ and $\bm \Delta$ onto the active eigenspaces. However, estimating the leading fraction $\theta^*$ can overcome a potential lack of solution-level information. Indeed,
\begin{align*}
    \frac{t^*}{T}=\theta^*+O(T^{-1}).
\end{align*}
The offset and integer-allocation error contribute only $O(T^{-1})$ to the compute fraction. This suggests estimating $\widehat{\theta}\approx \theta^*$ through the two bottleneck rates in the definition (\ref{def theta}) of $\theta^*$. We next describe one possible approach for the $\eh$-norm criterion, deployed in \cref{fig:linear}.

\textbf{Estimating the upstream bottleneck $\lambda_{i^*}$:} Let $\bm g:={\bm X^\top \bm y}/{n}=-\nabla R_{\operatorname{pre}}(\bm 0)=\bm H\ww$,
which is already needed for the first pretraining iteration. Since $\bm P_i \bm g=\lambda_i \bm P_i\ww$ and $\lambda_i>0$, the upstream bottleneck satisfies $\lambda_{i^*}=\min\{\lambda_i:\bm P_i \bm g\neq \bm 0\}$. Consequently, $\lambda_{i^*}$ can be estimated using a matrix-free Lanczos iteration (see, for instance, \cite{saad2011numerical}) applied to $\bm H$ and initialised with $\bm g$, without first computing $\ww$. Indeed, $\bm H^m \bm g=\sum_{i\in\mathcal I}\lambda_i^m\bm P_i \bm g$, so no inactive eigenspaces  enter the subspace explored by the Lanczos method. The smallest eigenvalue of $\bm H$ on the full subspace generated by these vectors $\bm H^m\bm g$ is exactly $\lambda_{i^*}$. 

\textbf{Estimating the downstream bottleneck $\lambda'_{j^*}$:} If $\bm P'_1\bm \Delta\neq \bm0$, as one expects, for instance, in a random-design setting, then $j^*=1$. Consequently, the relevant downstream eigenvalue is simply the smallest eigenvalue $\lambda'_1$ of $\eh$. When $d>n'$ we can simply use $\lambda_1'=\gamma'$, and otherwise it can be estimated, once again, by Lanczos, using a starting vector with a non-zero $\bm P'_1$ projection.

\textbf{Computational requirements:} Since they deploy matrix-free Lanczos iterations, these procedures use Hessian–vector products rather than explicit Hessian matrices. For example, $\bm H \bm v = \bm X^\top(\bm X\bm v)/n+\gamma \bm v$, which costs $O(nd)$ for a dense design matrix, with the analogous downstream product costing $O(n'd)$. For a fixed number of Lanczos iterations, estimating the allocation $\theta^*$ costs $O((n+n')d)$. For dense full-batch GD, the per-iteration costs are $\cp=\Theta(nd)$, $\cf=\Theta(n'd)$. As $B=LT$ and $L\geq\max\{\cp,\cf\}$, we have $(n+n')d=O(B/T)$.

In comparison, an exhaustive GD sweep over all $T+1$ allocations costs $\Theta(BT)$, even when the pretraining trajectory is shared. Alternatively, solving each ridge problem directly requires computing $\ww$ and $\ww'$, at a cost of $O((n+n')d^2)$ when $d\leq n,n'$ or $O((n^2+(n')^2)d)$ when $d\geq n,n'$. In either case, matrix-free Lanczos estimation is a more computationally efficient alternative.

 By \cref{thm:improvement}, for any fixed estimate $\widehat\theta\in(0,1)$, the feasible split nearest to $\widehat\theta T$ achieves improvement of $\exp(-2m(\widehat{\theta})T)$ over extreme allocations. Lanczos estimation error therefore affects the guaranteed exponential decay rate, rather than eliminating the guarantee, and this rate approaches its maximum as $\widehat\theta\to\theta^*$. This supports estimating the leading fraction $\theta^*$ without computing the ridge solutions or the offset~$s^*$. The accuracy of such an estimation requires a separate analysis that we leave as~future~work.

\section{Conclusion}
\label{sec:conclusion}

In this paper, we study compute allocation in a two-stage procedure for ridge regression: running $k$ GD iterations for pretraining, followed by fine-tuning with another $\ell$ GD iterations, under the budget constraint $\cp k +\cf\ell=B$. We characterise the optimal allocation $(k(t^*),\ell(t^*))$, showing that it is governed by the slowest upstream and downstream-visible bottleneck curvatures, with a correction term capturing their alignment. This characterisation is structural, requiring access to solution-level and spectral quantities, which in general can be as costly as solving the underlying optimisation problem itself (to compute the transfer update $\bm \Delta$). Algorithmically, we propose estimating the optimal allocation via matrix-free Lanczos iterations. This estimated interior allocation can be computed without access to solution-level information, and is more efficient than a checkpoint sweep or full spectral decompositions. To complement these results, we show that interior allocations are exponentially better (in $B$) than extreme splits.

Extending the theory to more general settings calls for new tools and perspectives. Our analysis relies on full-batch gradient descent, where the dynamics are linear and noise-free. Moving to stochastic optimisation would introduce gradient noise effects that interact non-trivially with the underlying geometry. More fundamentally, the ridge setting yields fixed Hessians and a static spectral decomposition, whereas in nonlinear models the representation, and thus the effective geometry, evolves during training. Understanding whether the same bottleneck-driven allocation principle persists in such feature-learning regimes, and how it can be approximated in an adaptive and computationally efficient way, remains an important direction for future work.

\section*{Acknowledgments}

Alex Buna was supported by EPSRC
through the StatML CDT programme [project reference EP/S023151/1].
Fanghui Liu was supported by the National Natural Science Foundation of China [grant number 12671631].
Patrick Rebeschini was supported by UK Research and Innovation (UKRI) under the UK government’s Horizon Europe funding guarantee [grant number EP/Y028333/1]

\newpage

\bibliographystyle{plainnat}
\bibliography{references}

\newpage
\appendix

\section{Additional notation}
\label{additionalnotation}
We introduce additional notation to aid with the proofs. When the eigendecomposition of a matrix $\bm A\in\mathbb{R}^{d\times d}$ is given by $(\lambda_i,\bm v_i)_{i=1}^d$ and $f$ is a scalar function, we write $f(\bm A)=\sum_{i=1}^d f(\lambda_i)\bm v_i\bm v_i^\top$.

Primarily, our analysis relies on the relaxation $f_{\tt M}:[0,T]\rightarrow[0,\infty)$, $f_{\tt M}(t)=\|\bm \Delta_t-\bm\Delta\|^2_{\tt M}$\ , which allows for differentiation. This can be expressed in a more convenient form using the two generating matrices
\begin{align*}
    \bm G:= -\frac{L}{\cp}\ln(\bm I-\eta\bm H),\quad\bm G':=-\frac{L}{\cf}\ln(\bm I-\eta'\bm H').
\end{align*}
Note that these are well-defined, symmetric, and positive definite, due to the step-size choice $0<\eta<\norm{\bm H}^{-1}$, $0<\eta'<\norm{\bm H'}^{-1}$\ . We work with these generating matrices, rather than the contractions $\bm I-\eta \bm H=\exp(-\cp \bm G/L)$, $\bm I-\eta' \bm H'=\exp(-\cf \bm G'/L)$, in order to allow for differentiation of exponentials with respect to $t$ in our subsequent analysis.

With this notation, note that the conclusions of \cref{lemma f(t) form} are
\begin{align*}
       \bm w_{k(t)}&=(\bm I-e^{-t \bm G}) \bm \ww\ ,\\
       \quad \bm w'_{\ell(t)} &= \bm \ww' + e^{-(T-t)\bm G'}(\bm w_{k(t)}- \bm \ww'),\\
        f_{\tt M}(t) &=\normg{(\bm I-e^{-(T-t)\bm G'})e^{-t\bm G} \bm \ww - e^{-(T-t) \bm G'}\bm{\Delta}}^2\ .
    \end{align*}

Additionally, recalling the basis-invariant eigendecompositions of $\bm H$ and $\bm H'$ as $(\lambda_i,\bm P_i)_{i=1}^r$ and $(\lambda'_j,\bm P'_j)_{j=1}^{r'}$, respectively, we denote the eigenvalues of $\bm G$ and $\bm G'$ as
\begin{align*}
    \mu_i:=-\frac{L}{\cp}\ln(1-\eta\lambda_i),\quad\mu_j':=-\frac{L}{\cf}\ln(1-\eta'\lambda_j'),
\end{align*}
where $1\leq i\leq r$ and $1\leq j\leq r'$. Note that the corresponding orthogonal eigenspace projectors of $\bm G$ and $\bm G'$ remain $\{\bm P_i\}_{i=1}^r$ and $\{\bm P_j'\}_{j=1}^{r'}$, respectively, and recall that we denote $i^*=\min \mathcal I$, $j^*=\min \mathcal{I}'$, potentially taking different values when $\texttt M=\eh$ or $\texttt M=\esp$. With this notation in place, the leading fraction becomes
\begin{align*}
    \theta^*=\frac{\mu'_{j^*}}{\mu_{i^*}+\mu'_{j^*}}.
\end{align*}

The expression of the offset term in both \cref{thm:localisationl2,thm:localisation} is
\begin{align*}
        s^*&:=\frac{1}{\mu_{i^*}+\mu'_{j^*}}\ln\left(\frac{(\mu'_{j^*}-\mu_{i^*})d^*+\sqrt{(\mu'_{j^*}-\mu_{i^*})^2(d^*)^2+4\mu_{i^*}\mu'_{j^*}a^*b^*}}{2\mu'_{j^*}b^*}\right),
\end{align*}
where $a^*:=\normg{\bm P_{i^*}\ww}^2>0$, $b^*:=\normg{\bm P'_{j^*}\bm{\Delta}}^2>0$, and $d^*:=\langle \bm P_{i^*}\ww, \bm P'_{j^*}\bm{\Delta}\rangle_{\tt M}$. Moreover, the convergence rate $\delta>0$ is given by
    \begin{align*}
        \delta&:=\min\left(\delta_1,\delta_2,\frac{\mu_{i^*}\mu'_1}{\mu_{i^*}+\mu'_{j^*}}\right),\\
        \text{where } \delta_1&:=\begin{cases}
            +\infty&\text{if } \mathcal{I}=\{i^*\},\\
            \underset{i\in \mathcal{I}\setminus\{i^*\}}{\min}\frac{(\mu_i-\mu_{i^*})\mu'_{j^*}}{\mu_{i^*}+\mu'_{j^*}}, &\text{otherwise},
        \end{cases}
        \,\,\, \delta_2:=\begin{cases}
            +\infty&\text{if } \mathcal{I}'=\{j^*\},\\
            \underset{j\in \mathcal{I}' \setminus\{j^*\}}{\min}\frac{\mu_{i^*}(\mu'_j-\mu'_{j^*})}{\mu_{i^*}+\mu'_{j^*}}, &\text{otherwise}.
        \end{cases}
    \end{align*}

    Additionally, the expression of the burn-in period in the (\ref{nkh}) condition of \cref{thm:localisation} (specialised for $\texttt{M}=\esp$) is given by:
    \begin{align*}
        K:=\begin{cases}
            0, &\text{if } \mathcal{I}=\{i^*\},\\
            \max\left(0,\left\lceil\frac{1}{\underset{i\in \mathcal{I}\setminus\{i^*\}}{\min}(\mu_i-\mu_{i^*})}\ln\left(\frac{4\sum_{i\in \mathcal{I} \setminus\{i^*\}}\norms{\bm{P}_i\ww}}{\norms{\bm P_{i^*}\ww}}\right)\right\rceil \right) &\text{otherwise.}
        \end{cases}
    \end{align*}
    Using the generator $\bm G$, the (\ref{nkh}) condition can be equivalently expressed as
    \begin{align*}
        \underset{t\in [K]}{\min}\norms{e^{-t \bm G}\ww}>0\ .
    \end{align*}

\section{Proof sketch of \cref{thm:localisation}}
\label{proofsketch}
In this section we give the proof sketch of \cref{thm:localisation} under the $\norms{\cdot}$ seminorm. The proof of \cref{thm:localisationl2} under the $\eh$-induced norm is similar with the exception of Step 5 below, so we omit the sketch of its proof in the main text. The main difference is that, by the invertibility of $e^{-t\bm G}$ and as $\ww\neq \bm 0$, $\smallnorm{e^{-t\bm G}\ww}_2>0$ always and we do not need to impose the (\ref{nkh}) assumption. 
The relevant notation is presented in \hyperref[additionalnotation]{Appendix~\ref*{additionalnotation}}, and full details and derivations can be found in \hyperref[app:localisation]{Appendix~\ref*{app:localisation}}. The proof is split into seven steps and it relies on analysing $f_{\esp}'$, the derivative of the interpolator $f_{\esp}:[0,T]\rightarrow[0,\infty)$. It requires identifying an approximate zero of $f_{\esp}'$, and developing a perturbative control of non-commuting dynamics at exponentially small scales.

\paragraph{Step 1: Decomposing $f'$:} First of all, we recenter $f$ around the limiting
\begin{align*}
    t_0:=\frac{\mu'_{j^*}}{\mu_{i^*}+\mu'_{j^*}}T\,,
\end{align*}
so that we mostly work with $f_{\esp}(t_0+s)$ with $s\in [-t_0,T-t_0]$. Under this reparametrisation, a bit of algebra shows that
\begin{align*}
    \frac{f_{\esp}'(t_0+s)}{2e^{-2m^* T}}=\Gamma(s)+\Xi(s),
\end{align*}
where we recall that $m^*=\mu_{i^*}\mu'_{j^*}/(\mu_{i^*}+\mu'_{j^*})$, and the two newly introduced functions of $s$ have the following roles: $\Gamma(s)$ gathers the three terms $a^*$, $b^*$ and $d^*$ (defined in \hyperref[additionalnotation]{Appendix~\ref*{additionalnotation}}) from the expansion terms of $f'_{\esp}$ that depend on $i^*$ and $j^*$ only, and hence are the relevant asymptotic bottlenecks, i.e.
\begin{align*}
    \Gamma(s)=\mu'_{j^*}\norms{\bm P'_{j^*}\bm{\Delta}}^2e^{2\mu'_{j^*} s}\!-\!\mu_{i^*}\norms{\bm P_{i^*}\ww}^2e^{-2\mu_{i^*} s}\!-\!(\mu'_{j^*}\!-\!\mu_{i^*}) \langle \bm P_{i^*}\ww, \bm P'_{j^*}\bm{\Delta}\rangle_{\esp} e^{(\mu'_{j^*}-\mu_{i^*})s};
\end{align*}
while $\Xi(s)$ encompasses the remaining terms in $f'$ that behave like `small' perturbations.

\paragraph{Step 2: Analysing $\Gamma(s)$:} It turns out that $\Gamma(s)$ has as its unique root the offset $s^*$. Thus, if $\Xi(s)$ has a small additive contribution, we would expect $f_{\esp}'(t_0+s^*)\approx 0$. 

Moreover, $\Gamma(s)$ is strictly increasing in a neighborhood $[s^*-S_0,s^*+S_0]$ centred at $s^*$, for some $S_0$ that does not depend on $T$, and we denote by
\begin{align*}
    m_{S_0}:=\underset{|s^*-s|\leq S_0}{\min}\Gamma'(s)>0.
\end{align*}
Moreover, as $\Gamma(s^*)=0$, we have that $\Gamma(s^*-S_0)<0$ and $\Gamma(s^*+S_0)>0$.

\paragraph{Step 3: Uniform bound for $\Xi(s)$ in any compact interval centred at $s^*$:} We show that for any $S>0$, the perturbation term $\Xi(s)$ and its derivative are `small' in the sense that
    \begin{align}
    \label{xi xi prime}
        \underset{|s-s^*|\leq S}{\sup}(|\Xi(s)|+|\Xi'(s)|)\leq C_Se^{-\delta T},
    \end{align}
where $C_S$ does not depend on $T$ and $\delta$ was introduced in the statement of \cref{thm:localisation}. Thus, we can expect $\Xi$ to not shift the zero of $f'$ far away from $t_0+s^*$.

\paragraph{Step 4: Existence, uniqueness and localisation of the zero of $\Gamma(s)+\Xi(s)$ inside $[s^*-S_0,s^*+S_0]$:} Since $S_0$ is independent of $T$, using \cref{xi xi prime} for $T$ large enough we can make $|\Xi'(s)|$ appropriately small on $[s^*-S_0,s^*+S_0]$ such that
\begin{align*}
    \Gamma'(s)+\Xi'(s)\geq m_{S_0}-|\Xi'(s)|\geq m_{S_0}/2>0
\end{align*}
So, if $\Gamma+\Xi$ has a zero in this interval, it is unique. 

And $\Gamma+\Xi$ indeed has a zero in $[s^*-S_0,s^*+S_0]$: As we argued in Step 2, $\Gamma(s^*-S_0)<0$ and by \cref{xi xi prime} we can make sure that the small additive contribution of $\Xi(s^*-S_0)$ does not change the sign of $\Gamma$. Similarly $(\Gamma+\Xi)(s^*+S_0)>0$, and by the Intermediate Value Theorem we get the existence of a zero, call it $z_T$.

To localise $z_T$, we can use the Mean Value Theorem on the interval $I\subseteq [s^*-S_0,s^*+S_0]$ with endpoints $z_T$ and $s^*$ and the fact that $\Gamma(s^*)=0$:
\begin{align*}
    C_{S_0}e^{-\delta T}&\geq |\Xi(s^*)|=|(\Xi+\Gamma)(s^*)-(\Xi+\Gamma)(z_T)|\\
    &\geq |s^*-z_T|\cdot \underset{\xi\in \mathcal{I}}{\inf}|\Gamma'(\xi)+\Xi'(\xi)|
        \geq |s^*-z_T|m_{S_0}/2,
\end{align*}
which, after defining $\bar t_T:=t_0+z_T$ (to translate it back to $t$-coordinates), rearranges as
\begin{align*}
        |\bar t_T-(t_0+s^*)|\leq \frac{2C_{S_0}}{m_{S_0}}e^{-\delta T}.
\end{align*}
This localises the continuous critical point $\bar t_T$. In the next steps we show that every discrete global minimiser is one of its adjacent integers.

\paragraph{Step 5: Proving the discrete global minimisers lie inside $t_0+[S_-,S_+]$ for some $S_-<0$, $S_-<s^*<S_+$:} To do so, we first upper-bound $f_{\esp}(\hat t_T)\leq C_0e^{-2m^* T}$, where $\hat t_T$ is the nearest integer to $t_0+s^*$. This can be easily done using the technical tools developed in Step 3. Next, we show that $f_{\esp}(t_0+s)\geq C_{S_-}e^{-2m^* T}$ for all $s\leq S_-$ and $f_{\esp}(t_0+s)\geq C_{S_+}e^{-2m^* T}$ for all $s\geq S_+$, in both cases working only with $s$ for which $t_0+s\in[T]$. Once we show how to pick $S_-<s^*<S_+$ such that $C_{S_{\pm}}>C_0$, we can conclude that the discrete global minimisers lie inside $t_0+[S_-,S_+]$. While the details are rather technical, we focus on the left-tail $s\leq S_-$ lower bound that uses assumption (\ref{nkh}):

One can show that the following holds for all $t\leq S_-+t_0$ for a carefully chosen $S_-<0$:
\begin{align*}
    \sqrt{f_{\esp}(t)}\geq\frac{3}{4}\norms{e^{-t \bm G}\ww}-e^{-\mu'_{j^*}(T-t)}\norms{\bm{\Delta}}.
\end{align*}
When $t\geq K$ introduced in \hyperref[additionalnotation]{Appendix~\ref*{additionalnotation}}, it can be shown that $\smallnorm{e^{-t \bm G}\ww}_{\esp}\geq 3e^{-\mu_{i^*}t}\smallnorm{\bm P_{i^*}\ww}_{\esp}/4$ and adjusting $S_-$ appropriately (dependent on the data, but independent of $T$), we can achieve the desired lower-bound on $f_{\esp}(t_0+s)$ for all $s\leq S_-$. However, such a uniformly positive lower-bound on $\smallnorm{e^{-t \bm G}\ww}_{\esp}$ cannot be guaranteed uniformly near $0$ (for example, at $t=0$, it could be the case that $\ww\in\operatorname{ker}(\esp)$ and the respective quantity is $0$), hence why at the very least we need $\smallnorm{e^{-t \bm G}\ww}_{\esp}$ to be bounded away from $0$ for $t\in [K]$.

Thus, all discrete global minimisers must be in the central interval $[S_-+t_0,S_++t_0]$.

\paragraph{Step 6: Sign of the derivative in the central interval:} In this step, we combine the results from Steps 4 and 5, together with a perturbative sign argument based on Step 3, to conclude that on the central interval the derivative changes sign at $\bar t_T$.

\paragraph{Step 7: Localisation of the discrete global minimisers:} Finally, we argue that each discrete minimiser is at one of the two integers adjacent to $\bar t_T$.

\section{Missing proofs from Section \ref{sec:localisation}}
\label{app:localisation}
\subsection{{ Proof of \hyperref[lemma: ridge connection]{Lemma~\ref*{lemma: ridge connection}}}}

Denote by $\bm u=\bm \Delta_t-\bm \Delta$, so that $\ww+\bm \Delta_t=\ww'+\bm u$ and so
\begin{align*}
    R_{\operatorname{fine}}(\ww+\bm \Delta_t)-R_{\operatorname{fine}}(\ww')&=R_{\operatorname{fine}}(\ww'+\bm u)-R_{\operatorname{fine}}(\ww')\\
    &=\frac{1}{2n'}\left(\norm{\bm X'(\ww'+\bm u)-\bm y'}^2-\norm{\bm X'\ww'-\bm y'}^2\right)\\
    &+\frac{\gamma'}{2}\left(\norm{\ww'+\bm u}^2-\norm{\ww'}^2\right)\\
    &=\frac{1}{n'}(\bm X'\ww'-\bm y')^\top \bm X'\bm u+\frac{1}{2n'}\norm{\bm X'\bm u}^2+\gamma'(\ww')^\top \bm u+\frac{\gamma'}{2}\norm{\bm u}^2\\
    &=\left(\frac{1}{n'}(\bm X')^\top(\bm X'\ww'-\bm y')+\gamma'\ww'\right)^\top \bm u+\frac{1}{2n'}\norm{\bm X'\bm u}^2+\frac{\gamma'}{2}\norm{\bm u}^2.
\end{align*}
Using the first order optimality condition for $\ww'$ with respect to $R_{\operatorname{fine}}(\cdot)$,
\begin{align*}
    \frac{1}{n'}(\bm X')^\top(\bm X'\ww'-\bm y')+\gamma'\ww'=0,
\end{align*}
and thus
\begin{align*}
    R_{\operatorname{fine}}(\ww+\bm \Delta_t)-R_{\operatorname{fine}}(\ww')&=\frac{1}{2n'}\norm{\bm X'\bm u}^2+\frac{\gamma'}{2}\norm{\bm u}^2=\frac{1}{2n'}\norm{\bm X'(\bm \Delta_t\!-\!\bm \Delta)}^2+\frac{\gamma'}{2}\norm{\bm \Delta_t\!-\!\bm \Delta}^2\\
    &=\frac{1}{2}\norms{\bm \Delta_t-\bm \Delta}^2+\frac{\gamma'}{2}\norm{\bm \Delta_t-\bm \Delta}^2=\frac{1}{2}\normh{\bm \Delta_t-\bm \Delta}^2,
\end{align*}
where the second to last equality follows as $\esp=((\bm X')^\top\bm X')/n'$ and $\eh=\esp+\gamma'\bm I$.

\subsection{{ Proof of \hyperref[lemma f(t) form]{Lemma~\ref*{lemma f(t) form}}}}
Define $\bm C:=\bm I-\eta\bm H$ and $\bm C':=\bm I-\eta' \bm H'$.
For the pretraining GD, we have for all $0\leq s\leq k(t)-1$,
\begin{align*}
    \bm w_{s+1} - \ww = \bm C(\bm w_s-\ww),
\end{align*}
and unfolding the recursion together with $\bm w_0=0$ we have
\begin{align*}
    \bm w_{k(t)}=(\bm I - \bm C^{k(t)})\ww.
\end{align*}

Similarly, for the fine-tuning GD, for all $0\leq s\leq \ell(t)-1$,
\begin{align*}
    \bm w_{s+1}'-\ww' = \bm C'(\bm w_s'-\ww'),
\end{align*}
and unfolding the recursion together with $\bm w'_{0}=\bm w_{k(t)}$ we arrive at
\begin{align*}
    \bm w'_{\ell(t)}=\ww'+(\bm C')^{\ell(t)}(\bm w_{k(t)}-\ww').
\end{align*}
 
 For the expression of $f_{\tt M}(t)=\smallnorm{\bm{\Delta}_t-\bm{\Delta}}^2_{\tt M}$, some algebra and the representation (\ref{eq:representation}) give:
\begin{align*}
    \bm{\Delta}_t-\bm{\Delta}&=(\bm w'_{\ell(t)}-\bm w_{k(t)})-\bm{\Delta} \\
    &=\ww'+(\bm C')^{\ell(t)}(\bm w_{k(t)}-\ww')-\bm w_{k(t)}-\bm{\Delta}\\
    &=((\bm C')^{\ell(t)}-\bm I)\bm w_{k(t)}+(\bm I-(\bm C')^{\ell(t)})\ww'-\bm{\Delta}\\
    &=((\bm C')^{\ell(t)}-\bm I)(\bm I-\bm C^{k(t)})\ww +(\bm I-(\bm C')^{\ell(t)})\ww'-\bm{\Delta}\\
    &=(\bm I-(\bm C')^{\ell(t)})\bm C^{k(t)}\ww + (\bm I-(\bm C')^{\ell(t)})(\ww'-\ww)-\bm{\Delta}\\
    &=(\bm I-(\bm C')^{\ell(t)})\bm C^{k(t)}\ww + (\bm I-(\bm C')^{\ell(t)})\bm{\Delta}-\bm{\Delta}\\
    &=(\bm I-(\bm C')^{\ell(t)})\bm C^{k(t)}\ww  -(\bm C')^{\ell(t)}\bm{\Delta}\,,
\end{align*}
which concludes the proof. \hfill$\blacksquare$

\subsection{{Proof of \cref{thm:pathologic}}} 
We first prove Case (i) for the $\norms{\cdot}$ seminorm, then Cases (ii) and (iii).

We start with $\mathcal{I'}=\emptyset$, which implies $\smallnorm{\bm P'_j\bm{\Delta}}_{\esp}^2=0$ for all $1\leq j\leq r'$. 
Using $\bm H' = \esp+\gamma' \bm I = \sum_{j=1}^{r'}\lambda'_j \bm P'_j$, $ \esp=\sum_{j=1}^{r'} \bm (\lambda'_j-\gamma')\bm P'_j$, we then obtain $\esp \bm P'_j=(\lambda'_j-\gamma')\bm P'_j$ via the orthogonality of $\{ \bm P'_j \}_{j=1}^{r'}$\ . Hence, for all $1\leq j\leq r'$, we have
\begin{align*}
    0=\norms{\bm P'_j\bm{\Delta}}^2=\bm{\Delta}^\top (\bm P'_j)^\top \esp \bm P'_j \bm{\Delta}=(\lambda'_j-\gamma')\bm{\Delta}^\top (\bm P'_j)^\top \bm P'_j\bm{\Delta}= (\lambda'_j-\gamma')\norm{\bm P'_j\bm{\Delta}}^2,
\end{align*}
so $\bm P'_j\bm{\Delta}=0$ for all $1\leq j\leq r'$ when $\lambda'_j>\gamma'$. Then,
\begin{align*}
    \norm{\esp\bm{\Delta}}^2=\bm{\Delta}^\top (\esp)^2\bm{\Delta}=\sum_{j:\lambda'_j=\gamma'}(\lambda'_j-\gamma')^2\norm{\bm P'_j\bm{\Delta}}^2+\sum_{j:\lambda'_j>\gamma'}(\lambda'_j-\gamma')^2\norm{\bm P'_j\bm{\Delta}}^2=0\,,
\end{align*}
so that $\esp\bm{\Delta}=0$. Then, using the fact that $\esp$ and $\bm H'$, hence $\esp$ and $e^{-(T-t)\bm G'}$ commute
\begin{align*}
    f_{\tt M}(t)&=\norms{(\bm I-e^{-(T-t)\bm G'})e^{-t \bm G}\ww - e^{-(T-t)\bm G'}\bm{\Delta}}^2\\
    &=\norms{(\bm I-e^{-(T-t)\bm G'})e^{-t \bm G}\ww}^2\\
    &-2((\bm I-e^{-(T-t)\bm G'})e^{-t \bm G}\ww)^\top e^{-(T-t)\bm G'}\esp\bm{\Delta}+(e^{-(T-t)\bm G'}\bm{\Delta})^\top e^{-(T-t)\bm G'}\esp\bm{\Delta}\\
    &=\norms{(\bm I-e^{-(T-t)\bm G'})e^{-t \bm G}\ww}^2.
\end{align*}
As $f_{\tt M}(T)=0$ and $f_{\tt M}\geq 0$, it is the case that $T\in\mathcal T^*$. This concludes (i).

On the other hand, if $\mathcal{I} =\emptyset$, then $\smallnorm{ \bm P_i\ww}_{\esp}= 0$ for all $1\leq i\leq r$, so that $(\esp)^{1/2} \bm P_i\ww= \bm 0$. Then
\begin{align*}
    \esp e^{-t \bm G}\ww=(\esp)^{1/2}\sum_{i=1}^{r}e^{-\mu_i t}(\esp)^{1/2} \bm P_i\ww= \bm 0\,.
\end{align*}
Again, using the fact that $\esp$ and $\bm I-e^{-(T-t)\bm G'}$ commute, we have $f_{\tt M}(t)=\smallnorm{e^{-(T-t)\bm G'}\bm{\Delta}}_{\esp}^2$\, which is minimised at $t^*=0$, proving (ii). 

These two cases combine to give (iii).

For the $\eh$ norm case, note that $\mathcal{I}'=\emptyset$ already implies that $e^{-(T-t)\bm G'}\bm{\Delta}=\bm 0$, which proves (i), and $\mathcal{I}=\emptyset$ implies that $e^{-t \bm G}\ww=\bm{0}$, proving (ii). These implications are due to the fact that $\eh$ is positive definite (unlike $\esp$), so $\normh{\bm v}=0$ if and only if $\bm v=\bm 0$.
\hfill $\blacksquare$

\subsection{{Proof of \cref{thm:localisationl2}}}

We split the proof into several steps. Throughout the Steps 1-6 of the proof, we will work with the interpolator of $f_{\eh}$, which we will define by overloading
\begin{align*}
    f_{\bm H'}:[0,T]\rightarrow[0,\infty),\quad f_{\bm H'}(t)=\normh{(\bm I-e^{-(T-t)\bm G'})e^{-t \bm G}\ww - e^{-(T-t)\bm G'}\bm{\bm{\Delta}}}^2.
\end{align*}
We will revert back to the original discrete $f_{\bm H'}:[T]\rightarrow[0,\infty)$ at the end, in Step 7.

\noindent\textbf{Step 1: Decomposing $f_{\eh}'$:} Define 
\begin{align*}
    \bm g(t):=(\bm I-e^{-(T-t)\bm G'})e^{-t \bm G}\ww - e^{-(T-t)\bm G'}\bm{\bm{\Delta}}\,.
\end{align*}
Recall that, as $\eh$ is positive definite, $i\notin\mathcal{I}$ or $j\notin\mathcal{I}'$ implies that $\bm P_i\ww=0$ or $\bm P'_j\bm{\Delta}=0$, respectively. Using the spectral transformations of $\bm G$ and $\bm G'$, we have
\begin{align*}
    \bm g(t)&=\sum_{i\in \mathcal{I}}e^{-\mu_i t}\bm P_i\ww-\sum_{i\in \mathcal{I}}\sum_{j=1}^{r'}e^{-\mu_i t-\mu'_j(T-t)}\bm P'_j\bm P_i\ww-\sum_{j\in \mathcal{I}'}e^{-\mu'_j(T-t)}\bm P'_j\bm{\Delta},\\
    \bm g'(t)&=-\sum_{i\in \mathcal{I}}\mu_ie^{-\mu_i t}\bm P_i\ww+\sum_{i\in \mathcal{I}}\sum_{j=1}^{r'}(\mu_i-\mu'_j)e^{-\mu_i t-\mu'_j(T-t)}\bm P'_j\bm P_i\ww-\sum_{j\in \mathcal{I}'}\mu'_je^{-\mu'_j(T-t)}\bm P'_j\bm{\Delta}.
\end{align*}
Further, note that $f_{\eh}'(t)=2g(t)^\top\bm H' g'(t)$. 

For ease of presentation, we use the following notations:
\begin{align*}
     c^*:=\frac{\mu'_{j^*}}{\mu_{i^*}+\mu'_{j^*}}, \quad t_0:=c^*T, \quad m^*:=\frac{\mu_{i^*}\mu'_{j^*}}{\mu_{i^*}+\mu'_{j^*}},\quad\kappa^*:=e^{-2m^*T}\,.
\end{align*}
Using the transformation $t=t_0+s$ with $s\in [-t_0,T-t_0]$, we can decompose
\begin{align*}
    \bm g(t_0+s)&=e^{-\mu_{i^*}(t_0+s)}\bm P_{i^*}\ww-e^{-\mu'_{j^*}(T-(t_0+s))}\bm{P}'_{j^*}\bm \Delta+\sum_{i\in \mathcal{I}\setminus\{i^*\}}e^{-\mu_i (t_0+s)}\bm P_i\ww\\
    &-\sum_{i\in \mathcal{I}}\sum_{j=1}^{r'}e^{-\mu_i (t_0+s)-\mu'_j(T-(t_0+s))}\bm P'_j\bm P_i\ww-\sum_{j\in\mathcal{I}'\setminus\{j^*\}}e^{-\mu'_j(T-(t_0+s))}\bm P'_j\bm\Delta\\
    &:=\underbrace{e^{-\mu_{i^*}(t_0+s)}\bm P_{i^*}\ww-e^{-\mu'_{j^*}(T-(t_0+s))}\bm P'_{j^*}\bm{\Delta}}_{\bm g_{\operatorname{lead}}}+\bm g_{\operatorname{rest}}(t_0+s),\\
    \bm g'(t_0+s)&=\underbrace{-\mu_{i^*}e^{-\mu_{i^*}(t_0+s)}\bm P_{i^*}\ww-\mu'_{j^*}e^{-\mu'_{j^*}(T-(t_0+s))}\bm P'_{j^*}\bm{\Delta}}_{\bm g_{\operatorname{lead}}'}+\bm g'_{\operatorname{rest}}(t_0+s)
\end{align*}
and further, taking the $\eh$-inner product between these two, we have
\begin{align*}
    \frac{f_{\eh}'(t_0+s)}{2}&=\gl(t_0+s)^\top \eh \glp(t_0+s)\\
    &+\underbrace{(\gr^\top \eh \glp+\gl^\top \eh\grp+\gr^\top \eh\grp)(t_0}_{:=\kappa^*\cdot \Xi}+s)\,,
\end{align*}
where we have introduced, for notation simplicity, the function $\Xi:[-t_0,T-t_0]\rightarrow\mathbb{R}$.

Now, note that $\sqrt{\kappa^*}=e^{-m^*T}=e^{-\mu_{i^*}t_0}=e^{-\mu'_{j^*}(T-t_0)}$, so that
\begin{align*}
    \Gamma(s)&:=\frac{\gl(t_0+s)^\top\eh\glp(t_0+s)}{\kappa^*}\\
    &=\mu'_{j^*}\normh{\bm P'_{j^*}\bm{\Delta}}^2e^{2\mu'_{j^*} s}\!-\!\mu_{i^*}\normh{\bm P_{i^*}\ww}^2e^{-2\mu_{i^*} s}\!-\!(\mu'_{j^*}\!-\!\mu_{i^*}) \ww^\top \bm P_{i^*}\eh\bm P'_{j^*}\bm{\Delta} e^{(\mu'_{j^*}-\mu_{i^*})s},
\end{align*}
and further we have decomposed $f'$ into (as we show below) a quadratic $\Gamma$ in the exponential space plus a perturbation term $\Xi$:
\begin{align}
\label{f' decomp2}
    \frac{f_{\eh}'(t_0+s)}{2\kappa^*}=\Gamma(s)+\Xi(s).
\end{align}

We next compute the single zero of $\Gamma$ and show that around this zero the function $\Gamma$ is increasing: 
    
\bigskip

    \noindent\textbf{Step 2: Analysing $\Gamma(s)$:} For brevity, we will use the notation
    \begin{align}
    \label{abd2}
        a^*=\normh{\bm P_{i^*}\ww}^2,\quad b^*:=\normh{\bm P'_{j^*}\bm{\Delta}}^2,\quad d^*:=\ww^\top \bm P_{i^*}\eh\bm P'_{j^*}\bm{\Delta}.
    \end{align}
    With $q:=e^{(\mu_{i^*}+\mu'_{j^*})s}>0$, we have
    \begin{align*}
        e^{2\mu_{i^*} s}\Gamma(s)=\mu'_{j^*}b^*q^2-(\mu'_{j^*}-\mu_{i^*})d^* q -\mu_{i^*} a^*.
    \end{align*}
    This is a quadratic in $q>0$. The product of the roots is negative, so the quadratic has exactly one positive root. Taking its $\log$ and dividing by $\mu_{i^*}+\mu'_{j^*}$\ we arrive at the unique zero of $\Gamma$:
    \begin{align*}
        s^*=\frac{1}{\mu_{i^*}+\mu'_{j^*}}\ln\left(\frac{(\mu'_{j^*}-\mu_{i^*})d^*+\sqrt{(\mu'_{j^*}-\mu_{i^*})^2(d^*)^2+4\mu_{i^*}\mu'_{j^*}a^*b^*}}{2\mu'_{j^*} b^*}\right).
    \end{align*}
    Next, using the fact that $\Gamma(s^*)=0 \iff (\mu'_{j^*}-\mu_{i^*})d^* e^{(\mu'_{j^*}-\mu_{i^*})s^*}=\mu'_{j^*}b^*e^{2\mu'_{j^*}s^*}-\mu_{i^*}a^*e^{-2\mu_{i^*}s^*}$, we have
    \begin{align*}
        \Gamma'(s^*)&=2(\mu'_{j^*})^2b^*e^{2\mu'_{j^*}s^*}+2(\mu_{i^*})^2a^*e^{-2\mu_{i^*}s^*}-(\mu'_{j^*}-\mu_{i^*})^2d^*e^{(\mu'_{j^*}-\mu_{i^*})s^*}\\
        &=2(\mu'_{j^*})^2b^*e^{2\mu'_{j^*}s^*}+2(\mu_{i^*})^2a^*e^{-2\mu_{i^*}s^*}-(\mu'_{j^*}-\mu_{i^*})(\mu'_{j^*}b^*e^{2\mu'_{j^*}s^*}-\mu_{i^*}a^*e^{-2\mu_{i^*}s^*})\\
        &=(\mu_{i^*}+\mu'_{j^*})(\mu'_{j^*}b^*e^{2\mu'_{j^*}s^*}+\mu_{i^*}a^*e^{-2\mu_{i^*}s^*})>0.
    \end{align*}
    So, there exists $S_0>0$ (independent of $T$) such that $m_{S_0}:=\min_{|s-s^*|\leq S_0}\Gamma'(s)>0$. In this interval $[s^*-S_0,s^*+S_0]$ the function $\Gamma$ is thus increasing. It is negative on $[s^*-S_0,s^*)$ and positive on $(s^*,s^*+S_0]$. Moreover, one can pick $T$ large enough to make sure $[t_0+s^*-S_0,t_0+s^*+S_0]\subseteq [0,T]$.

    Next, we show that the additive perturbation term $\Xi$ is small on any compact interval around $s^*$:

    \bigskip

    \noindent\textbf{Step 3: Uniform bound for the perturbation term $\Xi(s)$ in any compact $[s^*-S,s^*+S]$:} We show that for any $S>0$ (so, in particular, for the previous choice of $S_0$), there exist $C_S>0$ that may change from line to line, but stay independent of $T$ and just absorb other constants, and $\delta>0$ (also independent of $T$) such that for $T$ large enough
    \begin{align*}
        \underset{|s-s^*|\leq S}{\sup}(|\Xi(s)|+|\Xi'(s)|)\leq C_Se^{-\delta T}.
    \end{align*}
    Recalling the definition of $\Xi$ from \cref{f' decomp2} and using Cauchy-Schwarz,
    \begin{align}
    \label{|P| ridge2}
       |\Xi|&= \frac{1}{\kappa^*}|\bm g_{\operatorname{rest}}^\top\eh \bm g_{\operatorname{lead}}'+\bm g_{\operatorname{lead}}^\top\eh \bm g_{\operatorname{rest}}'+\bm g_{\operatorname{rest}}^\top \eh\bm g_{\operatorname{rest}}'| \nonumber\\
       &\leq \frac{1}{\kappa^*}\left(\normh{\bm g_{\operatorname{rest}}}\normh{\bm g'_{\operatorname{lead}}}+\normh{\bm g_{\operatorname{lead}}}\normh{\bm g'_{\operatorname{rest}}}+\normh{\bm g_{\operatorname{rest}}}\normh{\bm g'_{\operatorname{rest}}}\right).
    \end{align}
    We next bound each norm. For $s$ with $|s-s^*|\leq S$, as $|s|\leq |s^*|+S:=M_S$, we have
    \begin{align*}
        \left|e^{\pm\mu_i s}\right|=e^{\pm \mu_i s}\leq e^{\mu_i|s|}\leq e^{\mu_i M_S}\leq e^{\mu_r M_S}\leq e^{(\mu_{r}+\mu'_{r'})M_S}=:C_S,
    \end{align*}
    and similarly $|e^{\pm \mu'_j s}|\leq C_S$, $|e^{\pm(\mu'_j-\mu_i)s}|\leq C_S$.

    Recall that $\bm g_{\operatorname{lead}}(t_0+s)=e^{-\mu_{i^*}(t_0+s)}\bm P_{i^*}\ww-e^{-\mu'_{j^*}(T-(t_0+s))}\bm P'_{j^*}\bm{\Delta}$, $t_0=c^*T=\mu'_{j^*}T/(\mu_{i^*}+\mu'_{j^*})$ and the notations introduced in \cref{abd2}. Then, by a triangle inequality,
    \begin{align}
    \label{g lead2}
            \normh{\bm g_{\operatorname{lead}}(t_0+s)}&\leq e^{-\mu_{i^*} t_0}e^{-\mu_{i^*} s}\sqrt{a^*}+e^{-\mu'_{j^*}(T-t_0)}e^{\mu'_{j^*} s}\sqrt{b^*}\nonumber\\
            &=e^{-m^* T}e^{-\mu_{i^*} s}\sqrt{a^*}+e^{-m^* T}e^{\mu'_{j^*} s}\sqrt{b^*}\leq C_S e^{-m^* T}.
    \end{align}
     A similar computation (by letting $C_S$ absorb $\mu_{i^*}$ and $\mu'_{j^*}$) gives $\normh{\bm g'_{\operatorname{lead}}(t_0+s)}\leq C_S e^{-m^* T}$.

    By the definition of $\mu_{i^*}$ and $\mu'_{j^*}$\ , $\delta_1:=\underset{i\in \mathcal{I}\setminus\{i^*\}}{\min}(c^*(\mu_i-\mu_{i^*}))>0$ (taken to be $+\infty$ if $\mathcal{I}=\{i^*\}$), $\delta_2:=\underset{j\in \mathcal I'\setminus\{j^*\}}{\min}((1-c^*)(\mu'_j-\mu'_{j^*}))>0$ (taken to be $+\infty$ if $\mathcal{I}'=\{j^*\}$). Additionally, as $\bm H'$ is positive definite, $(1-c^*)\mu'_1>0$. We take $\delta:=\min(\delta_1,\delta_2,(1-c^*)\mu'_1)>0$.

     Next, we deal with bounding $\bm g_{\operatorname{rest}}$ and its derivative. However, for the derivative, we can simply absorb the eigenvalues into the constants, the analysis is otherwise identical. There are three types of contribution in it:

     \begin{itemize}
         \item From the first summation, terms with $i\neq i^*$: using the fact that $\mu_{i^*}c^*=m^*$, these terms can be bounded as
         \begin{align*}
             \normh{e^{-\mu_i(t_0+s)}\bm P_i\ww}&=e^{-\mu_i t_0}e^{-\mu_i s}\normh{\bm P_i\ww }\leq e^{-\mu_ic^*T}C_S\normh{\bm P_i\ww}\\
             &=e^{-(\mu_i-\mu_{i^*})c^*T}e^{-\mu_{i^*} c^* T}C_S\normh{\bm P_i\ww}\\
             &\leq C_Se^{-{\delta} T}e^{-m^* T}\normh{\bm P_i\ww}=C_Se^{-(m^*+{\delta})T}.
         \end{align*}
         where the constant (independent of $T$) $C_S$ absorbed $\normh{\bm P_i\ww}$.
         \item Similarly, from the third summation, the terms with $j\neq j^*$ can be bounded as
         \begin{align*}
             \normh{e^{-\mu'_j(T-(t_0+s))}\bm P'_j\bm{\Delta}}\leq C_Se^{-(m^*+{\delta})T}. 
         \end{align*}
         \item For the second summation over both $i\in I$ and $j\in \{1,\ldots,r'\}$, the contribution of each pair $(i,j)$ is:
         \begin{align*}
             \normh{e^{-\mu_i(t_0+s)-\mu'_j(T-(t_0+s))}\bm P'_j\bm P_i\ww}\leq C_Se^{-(\mu_ic^*+\mu'_j(1-c^*))T}.
         \end{align*}
         Now, note that 
         \begin{align*}
             \mu_ic^*+\mu'_j(1-c^*)&\geq \mu_{i^*}c^*+\mu'_1(1-c^*)=m^*+\mu'_1(1-c^*)\geq m^*+{\delta},
         \end{align*}
         so that the contribution for each pair $(i,j)$ is again at most $C_Se^{-({\delta}+m^*)T}$.
     \end{itemize}
     Putting everything together and absorbing the sum of all constants into a single constant $C_S$ (which depends on $r'$ and $|\mathcal{I}|$), we get that for any $s\in [s^*-S,s^*+S]$,
     \begin{align}
     \label{g rest2}
         \normh{\bm g_{\operatorname{rest}}(t_0+s)}\leq C_S e^{-(m^*+{\delta})T}\quad\text{and}\quad \normh{\bm g'_{\operatorname{rest}}(t_0+s)}\leq C_S e^{-(m^*+{\delta})T},
     \end{align}
     for some appropriate (independent of $T$, but containing $|\mathcal{I}|$ and $r'$) constant $C_S$.

     Using \cref{|P| ridge2}, we get that
     \begin{align*}
         |\Xi(s)|\leq \frac{C_S}{\kappa^*}(e^{-(2m^*+{\delta})T}+e^{-(2m^*+{\delta})T}+e^{-2(m^*+{\delta})T})\leq C_Se^{-{\delta} T}. 
     \end{align*}

     The bound for $\Xi'(s)$ is almost identical, with the only difference being that
     \begin{align*}
         \kappa^* \Xi'=2(\bm g'_{\operatorname{rest}})^\top \eh\bm g'_{\operatorname{lead}}+\bm g_{\operatorname{rest}}^\top\eh \bm g''_{\operatorname{lead}}+(\bm g''_{\operatorname{rest}})^\top \eh\bm g_{\operatorname{lead}}+\normh{\bm g'_{\operatorname{rest}}}^2+\bm g_{\operatorname{rest}}^\top \eh\bm g''_{\operatorname{rest}},
     \end{align*}
     and now the bounds for the second derivatives are (up to constants independent of $T$) identical.

     Given this bound on the perturbation term $\Xi$ on any compact interval around $s^*$, we show that it cannot perturb the zero of $\Gamma(s)$ by too much:

     \bigskip
     \noindent\textbf{Step 4: Existence, uniqueness and localisation of the zero of $\Gamma(s)+\Xi(s)$ inside $[s^*-S_0,s^*+S_0]$:} First, we show that $\Gamma(s)+\Xi(s)$ has a unique zero in $(s^*-S_0,s^*+S_0)$ for $T$ large enough. Start by defining
     \begin{align*}
         \eta_{S_0}:=\min\{-\Gamma(s^*-S_0),\Gamma(s^*+S_0)\}
     \end{align*}
     and note that $\eta_{S_0}>0$, as the derivative in $[s^*-S_0,s^*+S_0]$ is positive by the choice of $S_0$ and $\Gamma(s^*)=0$. Moreover, pick $T_0$ large enough such that, for all $T\geq T_0$,
     \begin{align*}
         C_{S_0}e^{-\delta T}\leq \min\{\eta_{S_0}/2,m_{S_0}/2\},
     \end{align*}
     where we recall that $m_{S_0}:=\min_{|s-s^*|\leq S_0}\Gamma'(s)>0$ and $C_{S_0}$ is the constant from the previous Step 3 when instantiating $S:=S_0$. This is possible as $S_0$ is independent of $T$. Then, for all $s\in [s^*-S_0,s^*+S_0]$, $|\Xi(s)|\leq \eta_{S_0}/2$ and $|\Xi'(s)|\leq m_{S_0}/2$. This implies that on that interval
     \begin{align}
     \label{Q'+P' ridge2}
         \Gamma'(s)+\Xi'(s)\geq m_{S_0}-|\Xi'(s)|\geq m_{S_0}/2>0,
     \end{align}
     so if there is a zero in this interval, it is unique. And there is indeed a zero: at the endpoints
     \begin{align*}
         &\Gamma(s^*-S_0)+\Xi(s^*-S_0)\leq -\eta_{S_0}+\eta_{S_0}/2=-\eta_{S_0}/2<0 \quad \text{and}\quad \\ &\Gamma(s^*+S_0)+\Xi(s^* + S_0)\geq \eta_{S_0}-\eta_{S_0}/2=\eta_{S_0}/2>0.
     \end{align*}
    and the existence follows by the Intermediate Value Theorem. Let $z_T$ be this unique zero. Since $f'(s+t_0)=2\kappa^* (\Gamma(s)+\Xi(s))$ is negative at the left end of the interval, $0$ at $z_T$, positive at the right end, it follows that $z_T+t_0$ is the minimum of $f$ in this interval.
    
    Further, note that $z_T$ satisfies $(\Gamma+\Xi)(z_T)=0$  and recall that $\Gamma(s^*)=0$. Then, by the Mean Value Theorem on the interval $I\subseteq [s^*-S_0,s^*+S_0]$ with endpoints $z_T$ and $s^*$,
    \begin{align*}
        |\Xi(s^*)|&=|(\Xi+\Gamma)(s^*)-(\Xi+\Gamma)(z_T)|\geq |s^*-z_T|\underset{\xi\in \mathcal{I}}{\inf}|\Gamma'(\xi)+\Xi'(\xi)|\\
        &\geq |s^*-z_T|\underset{\xi\in [s^*-S_0,s^*+S_0]}{\inf}|\Gamma'(\xi)+\Xi'(\xi)|\geq |s^*-z_T|m_{S_0}/2,
    \end{align*}
    where the last inequality follows from \cref{Q'+P' ridge2}. Next, recall that $|\Xi(s^*)|\leq C_{S_0}e^{-\delta T}$, so that
    \begin{align*}
        |z_T-s^*|\leq \frac{2C_{S_0}}{m_{S_0}}e^{-\delta T}.
    \end{align*}
    Defining $\bar{t}_T:=t_0+z_T$, we arrive at
    \begin{align*}
        |\bar{t}_T-(t_0+s^*)|\leq \frac{2C_{S_0}}{m_{S_0}}e^{-\delta T},
    \end{align*}
    and it remains to show that this is indeed the global minimum of the interpolator $f_{\eh}$, having previously argued it is a minimum in $[t_0+s^*-{S_0},t_0+s^*+{S_0}]$. 
    
    We do this by showing that:
    \begin{itemize}[leftmargin=*]
        \item (Step 5) the global minimum must be in some (other) compact interval $t_0+[S_-,S_+]$ around $t_0+s^*$;
        \item (Step 6) The function $s\mapsto f_{\eh}'(t_0\!+\!s)\!=\!2\kappa^*(\Gamma(s)\!+\!\Xi(s))$ changes sign exactly once on $[S_-,S_+]$.
    \end{itemize}

    \bigskip
    \noindent\textbf{Step 5: Proving the global minimum lies inside $t_0+[S_-,S_+]$ for some $S_-<0$, $S_-<s^*<S_+$:} We show that the minimum of $f$ has to lie in an interval containing $t_0+s^*$ and not close to the endpoints $0$ or $T$ of the domain of $f$. To do so, note that $\underset{t\in[0,T]}{\min} f_{\eh}(t)\leq f_{\eh}(t_0+s^*)$, which we estimate, and then argue that around $0$ and $T$ the function $f$ takes a higher value.
    
    \textbf{Upper-bounding $f_{\eh}(t_0+s^*)$:} Recall that $f_{\eh}(t_0+s^*)=\normh{\bm g(t_0+s^*)}^2$. For later use in Step 7, define
    \begin{align*}
        \hat t_T\in\arg\min_{q\in [T]}|q-(t_0+s^*)|,\quad \hat s_T:=\hat t_T-t_0.
    \end{align*}
    For a sufficiently large $T$, we have $|\hat{s}_T-s^*|\leq 1/2$.
    
    Now, we can apply the bounds derived in Step 3. To be more precise, consider $S=1$, then, according to the triangle inequality, \cref{g lead2} and \cref{g rest2}:
    \begin{align*}
        \sup_{|s-s^*|\leq 1}f_{\eh}(t_0+s)&=\sup_{|s-s^*|\leq 1}{\normh{\bm g(t_0+s)}^2}\\ &\leq \sup_{|s-s^*|\leq 1} (\normh{\gl(t_0+s)}+\normh{\gr(t_0+s)})^2\\
        &\leq C_1^2(e^{-m^* T}+e^{-(m^*+{\delta})T})^2
        \leq 2C_1^2 e^{-2m^* T}=2C_1^2 \kappa^*.
    \end{align*}
    Importantly, the constant does not depend on $T$. Since both $s:=s^*$ and $s:=\hat s_T$ satisfy $|s-s^*|\leq 1$, this inequality gives both
    \begin{align}
        \label{step 5}
        f_{\eh}(t_0+s^*)\leq 2C_1^2\kappa^*,
    \end{align}
    and
    \begin{align}
        \label{step 5 discrete}
        f_{\eh}(\hat t_T)\leq 2C_1^2\kappa^*.
    \end{align}

    \textbf{Left-tail lower bound:} We argue that for a certain $S_-$ (independent of $T$) and all $T$ large enough, i.e. $T\geq T_-(S_-)$ depending on $S_-$, we have for all $s\leq S_-$
    \begin{align*}
        f_{\eh}(t_0+s)\geq \frac{\gamma'\norm{\bm P_{i^*}\ww}^2e^{-2\mu_{i^*}S_-}}{4}\kappa^*.
    \end{align*}

    Since $\eh=\esp+\gamma'\bm I\succeq \gamma' I$ and $f_{\eh}(t_0+s)=\normh{\bm g(t)}^2\geq \gamma'\norm{\bm g(t)}^2$, it's in fact enough to show that
    \begin{align}
    \label{left goal2}
        \norm{\bm g(t_0+s)}^2\geq \frac{\norm{\bm P_{i^*}\ww}^2e^{-2\mu_{i^*}S_-}}{4}\kappa^*.
    \end{align}

    By the triangle inequality applied twice and the definition of the operator norm,
    \begin{align*}
        \norm{\bm g(t)}&\geq \norm{(\bm I-e^{-(T-t)\bm G'})e^{-t\bm G}\ww}-\norm{e^{-(T-t)\bm G'}\bm{\Delta}}\\
        &\geq \left(\norm{e^{-t\bm G}\ww}-\norm{e^{-(T-t)\bm G'}e^{-t \bm G}\ww}\right)-\norm{e^{-(T-t)\bm G'}\bm{\Delta}}\\
        &\geq  \left(\norm{e^{-t\bm G}\ww}-\norm{e^{-(T-t)\bm G'}}\norm{e^{-t\bm G}\ww}\right)-\norm{e^{-(T-t)\bm G'}\bm{\Delta}}\\ 
        &= \left(1-\norm{e^{-(T-t)\bm G'}}\right)\norm{e^{-t\bm G}\ww}-\norm{e^{-(T-t)\bm G'}\bm{\Delta}}.
    \end{align*}
     Now, consider $S_-<s^*$ to be fixed later and recall that we translated $t=t_0+s$. For $s\leq S_-$, $T-(t_0+s)\geq (1-c^*)T-S_-\rightarrow\infty$ as $T\rightarrow\infty$, so we can pick $T$ large enough (and dependent on $S_-$) such that
    \begin{align*}
        \norm{e^{-(T-t)\bm G'}}\leq\frac{1}{4}\quad \text{for all } s\leq S_-,
    \end{align*}
    and then we get that for any $s\leq S_-$,
    \begin{align}
    \label{g(t_0+s) left2}
        \norm{\bm g(t_0+s)}\geq \frac{3}{4}\norm{e^{-(t_0+s)\bm G}\ww}-\norm{e^{-(T-(t_0+s))\bm G'}\bm{\Delta}}.
    \end{align}
    We now bound the two norms:
    \begin{itemize}
        \item Because $\bm P_i$ are orthogonal projectors, the vectors $\bm P_i\ww$ are orthogonal and hence
        \begin{align*}
            \norm{e^{-(t_0+s)\bm G}\ww}^2&=\norm{\sum_{i\in \mathcal{I}}e^{-\mu_i (t_0+s)}\bm P_i\ww}^2=\sum_{i\in \mathcal I}e^{-2\mu_i (t_0+s)}\norm{\bm P_i\ww}^2\\
            &\geq e^{-2\mu_{i^*} (t_0+s)}\norm{\bm P_{i^*}\ww}^2,
        \end{align*}
        so that
        \begin{align}
        \label{Bw2}
             \norm{e^{-(t_0+s)\bm G}\ww}\geq e^{-\mu_{i^*} (t_0+s)}\norm{\bm P_{i^*}\ww}=e^{-m^*T}e^{-\mu_{i^*} s}\norm{\bm P_{i^*}\ww}
        \end{align}
        \item Since $\bm P'_j\bm\Delta=0$ for $j\notin\mathcal{I}'$ and $\mu'_{j^*}$ is the smallest eigenvalue over $\mathcal{I}'$, 
        \begin{align*}
            \norm{e^{-(T-(t_0+s))\bm G'}\bm{\Delta}}\leq e^{-\mu'_{j^*}(T-t)}\norm{\bm{\Delta}}
        \end{align*}
    \end{itemize}
    Then, for all $s\leq S_-$,
    \begin{align}
    \label{AB ratio2}
        \frac{\norm{e^{-(T-(t_0+s))\bm G'}\bm{\Delta}}}{\norm{e^{-(t_0+s)\bm G}\ww}}\leq e^{s(\mu'_{j^*}+\mu_{i^*})}\frac{\norm{\bm{\Delta}}}{\norm{\bm P_{i^*}\ww}}\leq e^{S_-(\mu'_{j^*}+\mu_{i^*})}\frac{\norm{\bm{\Delta}}}{\norm{\bm P_{i^*}\ww}}.
    \end{align}
    Hence, we can choose $S_-<s^*$ negative enough so that the right-hand side of \cref{AB ratio2} is $\leq 1/4$, from which
    \begin{align*}
        {\norm{e^{-(T-(t_0+s))\bm G'}\bm{\Delta}}}\leq \frac{1}{4}\norm{e^{-(t_0+s)\bm G}\ww}\quad\text{for all } s\leq S_-.
    \end{align*}
    Combining this with \cref{g(t_0+s) left2} and \cref{Bw2}, we get
    \begin{align*}
        \norm{\bm g(t_0+s)}\geq \frac{1}{2}\norm{e^{-(t_0+s)\bm G}\ww}\geq \frac{1}{2}e^{-m^* T}e^{-\mu_{i^*} s}\norm{\bm P_{i^*}\ww},
    \end{align*}
    and, finally, as $\kappa^*=e^{-2m^* T}$ and $f_{\eh}(t_0+s)=\normh{\bm g(t_0+s)}^2$, we arrive at \cref{left goal2} by squaring the above.
    
    \textbf{Right-tail lower-bound:} We argue that for a certain $S_+$ independent of $T$, we have for all $s\geq S_+$
    
    \begin{align*}
        f_{\eh}(t_0+s)\geq \frac{9}{16}\gamma'\norm{\bm P'_{j^*}\bm{\Delta}}^2e^{2\mu'_{j^*} S_+}\kappa^*.
    \end{align*}

    Similarly to the left-tail lower bound, it's in fact enough to show
    
    \begin{align}
    \label{right goal2}
        \norm{\bm g(t_0+s)}^2\geq \frac{9}{16}\norm{\bm P'_{j^*}\bm{\Delta}}^2e^{2\mu'_{j^*} S_+}\kappa^*.
    \end{align}
    For this, by the triangle inequality and the definition of the spectral norm,
    \begin{align*}
        \norm{\bm g(t)}&\geq  \norm{e^{-(T-t)\bm G'}\bm{\Delta}}-\norm{(\bm I-e^{-(T-t)\bm G'})e^{-t\bm G}\ww}\\
        &\geq \norm{e^{-(T-t)\bm G'}\bm{\Delta}}-\norm{\bm I-e^{-(T-t)\bm G'}}\norm{e^{-t\bm G}\ww}\\
        &\geq \norm{e^{-(T-t)\bm G'}\bm{\Delta}}-\norm{e^{-t\bm G}\ww}.
    \end{align*}
    We now bound the two norms:
    \begin{itemize}
        \item Since $e^{-\mu'_{j^*}(T-t)}\bm P_{j^*}'\bm{\Delta}$ is an orthogonal component of $e^{-(T-t)\bm G'}\bm{\Delta}$:
        \begin{align}
        \label{obs lower2}
            \norm{e^{-(T-t)\bm G'}\bm{\Delta}}\geq e^{-\mu'_{j^*}(T-t)}\norm{\bm P'_{j^*}\bm{\Delta}}.
        \end{align}
        \item Since $\bm P_i\bm\ww=0$ for $i\notin\mathcal{I}$ and $\mu_{i^*}$ is the smallest eigenvalue over $\mathcal{I}$, $$\norm{e^{-t\bm G}\ww}\leq e^{-t\mu_{i^*}}\norm{\ww}.$$
    \end{itemize}
    Putting these together and recalling that $t=t_0+s$ with $\mu_{i^*} t_0=\mu'_{j^*}(T-t_0)$, we obtain that for all $s\geq S_+$,
    \begin{align}
    \label{quotient upper2}
        \frac{\norm{e^{-t\bm G}\ww}}{\norm{e^{-(T-t)\bm G'}\bm{\Delta}}}\leq \frac{e^{-\mu_{i^*} t}\norm{\ww}}{e^{-\mu'_{j^*}(T-t)}\norm{\bm P'_{j^*}\bm{\Delta}}}=\frac{e^{-(\mu_{i^*}+\mu'_{j^*})s}\norm{\ww}}{\norm{\bm P'_{j^*}\bm{\Delta}}}\leq \frac{e^{-(\mu_{i^*}+\mu'_{j^*})S_+}\norm{\ww}}{\norm{\bm P'_{j^*}\bm{\Delta}}}.
    \end{align}
    Consequently, we can pick $S_+>s^*$ large enough so that the right-hand side of \cref{quotient upper2} is $\leq 1/4$. Then, $\norm{e^{-t\bm G}\ww}\leq \norm{e^{-(T-t)\bm G'}\bm{\Delta}}/4$ so that, together with \cref{obs lower2},
    \begin{align*}
        \norm{\bm g(t)}\geq \frac{3}{4}\norm{e^{-(T-t)\bm G'}\bm{\Delta}}\geq \frac{3}{4}e^{-\mu'_{j^*}(T-t)}\norm{\bm P'_{j^*}\bm{\Delta}}.
    \end{align*}
    Squaring this and using $f_{\eh}(t)\!=\!\normh{\bm g(t)}^2\!\geq\! \gamma'\norm{\bm g(t)}^2$ and $e^{-2\mu'_{j^*} (T-t)}\!=\!\kappa^*e^{2\mu'_{j^*} s}$, we arrive at \cref{right goal2}.

    \textbf{Putting everything together:}
    By decreasing $S_-$ and increasing $S_+$, if necessary, we have proved the following: there are $S_-<s^*-1<s^*+1<S_+$ independent of $T$ and $T$ large enough such that
    \begin{itemize}
        \item $f_{\eh}(t_0+s^*)\leq 2C_1^2\kappa^*$;
        \item $f_{\eh}(t_0+s)\geq (\gamma'\norm{\bm P_{i^*}\ww}^2 e^{-2\mu_{i^*} S_-}/4) \kappa^*$ for all $s\leq S_-$;
        \item $f_{\eh}(t_0+s)\geq (9\gamma'\norm{\bm P'_{j^*}\bm{\Delta}}^2 e^{2\mu'_{j^*} S_+}/16)\kappa^*$ for all $s\geq S_+$.
    \end{itemize}
    Recall that $C_1$ is a constant independent of $S_{\pm}$ or $T$. We can accordingly increase $S_+$ and decrease (and make negative) $S_-$ such that
    \begin{align*}
        \gamma'\norm{\bm P_{i^*}\ww}^2 e^{-2\mu_{i^*} S_-}/4\geq 3 C_1^2 \quad \text{and} \quad 9\gamma'\norm{\bm P'_{j^*}\bm{\Delta}}^2 e^{2\mu'_{j^*} S_+}/16\geq 3C_1^2.
    \end{align*}
    So, for these choices of $S_-, S_+$ and sufficiently large $T$ depending on the chosen $S_{-},S_+$, we have
    \begin{align*}
        \underset{S_-\leq s\leq S_+}{\min}f_{\eh}(t_0+s)&\leq f_{\eh}(t_0+s^*)\leq 2C_1^2\kappa^*\leq f_{\eh}(t_0+s) &\text{for all } s\leq S_-,\\
        \underset{S_-\leq s\leq S_+}{\min}f_{\eh}(t_0+s)&\leq f_{\eh}(t_0+s) &\text{for all } s\geq S_+.
    \end{align*}
    Accordingly, any global minimiser of $f_{\eh}$ must be of the form $t_0+s$ with $s\in [S_-,S_+]$. Note that, as $t_0=\mu'_{j^*}T/(\mu_{i^*}+\mu'_{j^*})$, we can enlarge $T$ such that this minimiser is indeed in $[0,T]$, i.e. $t_0+S_-\geq 0$, $t_0+ S_+\leq T$.

    Moreover, note that the discrete $\hat t_T\in[S_-+t_0,S_++t_0]$, and thus, using \cref{step 5 discrete},
    \begin{align*}
        \underset{\substack{q\in [T]\\t_0+S_-\leq q\leq t_0+S_+}}{\min} f_{\eh}(q)\leq f_{\eh}(\hat t_T)\leq 2C_1^2\kappa^*,
    \end{align*}
    whereas from the tail bounds we have that for every $q\in[T]$ outside $[t_0+S_-,t_0+S_+]$,
    \begin{align*}
        f_{\eh}(q)\geq 3C_1^2\kappa^*.
    \end{align*}
    Therefore, the set of global minimisers of the discrete $f_{\eh}$ satisfies
    \begin{align}
        \label{discrete global min} \mathcal T^*\subseteq [T]\cap [t_0+S_-,t_0+S_+].
    \end{align}

    \bigskip

    \noindent\textbf{Step 6: Uniqueness of the global minimiser:} Recall that we are still working with the interpolator $f_{\eh}:[0,T]\rightarrow[0,\infty)$. Also, recall the definition of $S_0$ from Step 2. Now, decrease $S_-$ and increase $S_+$, if needed, such that $S_-<s^*-S_0-1<s^*+S_0+1<S_+$, and furthermore consider $R>0$ for which $[S_-,S_+]\subseteq[-R,R]$. Additionally, we can increase $T$ until $0<t_0-R<t_0+R<T$, ensuring thus that any global minimiser in $[S_-+t_0,S_++t_0]$ is an interior point of the domain $[0,T]$.
    
    By Step 3, we have
    \begin{align*}
        \underset{-R\leq s\leq R}{\sup}|\Xi(s)|\leq C_R e^{-\delta T}. 
    \end{align*}
    Note that, as $\Gamma(s)=0$ has a unique solution at $s^*$, on $[-R,R]\setminus [s^*-S_0,s^*+S_0]$ the function $\Gamma$ does not have a zero and moreover
    \begin{align*}
        {\min}\{|\Gamma(s)|:s\in [-R,s^*-S_0]\cup [s^*+S_0,R]\}>0,
    \end{align*}
    which is independent of $T$. So, we can take $T$ large enough so that
    \begin{align*}
        \underset{-R\leq s\leq R}{\sup}|\Xi(s)|\leq\frac{1}{2}{\min}\{|\Gamma(s)|:s\in [-R,s^*-S_0]\cup [s^*+S_0,R]\}.
    \end{align*}
    Accordingly, for every $s\in[-R,R]\setminus [s^*-S_0,s^*+S_0]$,
    \begin{align*}
        \text{sign}(\Xi(s)+\Gamma(s))=\text{sign}(\Gamma(s)),
    \end{align*}
    as $|\Xi|\leq |\Gamma|/2$ implies that addition of $\Xi$ cannot change the sign of $\Gamma$. But, as we showed in Step 2, $s^*$ is the unique root of $\Gamma$ and $\Gamma(s)<0$ for $s< s^*$, $\Gamma(s)>0$ for $s>s^*$, so there is no root of $\Xi+\Gamma$ in $[-R,R]\setminus [s^*-S_0,s^*+S_0]$.
    
    By Step 4, $z_T$ is the unique root of $\Gamma+\Xi$ in $[s^*-S_0,s^*+S_0]$, while the above argument shows that there are no roots in $[S_-,S_+]\setminus [s^*-S_0,s^*+S_0]$. Hence, $t_0+z_T$ is the unique stationary point of $f_{\eh}$ in $[S_-+t_0,S_++t_0]$. But, by Step 5, every global minimiser lies within this interval and, since the interval is a subset of $(0,T)$, every such minimiser must be stationary. Thus, $\bar t_T=t_0+z_T$ is the unique global minimiser of the interpolator $f_{\eh}$.

    \bigskip
    \noindent\textbf{Step 7: Localisation of the discrete global minimisers.} By Step 5 \cref{discrete global min}, every $t^*\in\mathcal{T}^*$ lies in $[t_0+S_-,t_0+S_+]$. From Steps 4 and 6, the interpolator $f_{\eh}$ is strictly decreasing on $[t_0+S_-,\bar t_T]$, and strictly increasing on $[\bar t_T,t_0+S_+]$. Thus, its minimum over integer points in this interval can occur only at the two integers adjacent to $\bar t_T$, i.e.
    \begin{align*}
        \mathcal T^*\subseteq \{\lfloor \bar t_T\rfloor, \lceil \bar t_T\rceil\}.
    \end{align*}
    If $\bar t_T$ is an integer, these two points coincide. Hence, $\mathcal T^*$ contains at most two elements. Moreover, for every $t^*\in\mathcal T^*$, by the triangle inequality
    \begin{align*}
        |t^*-(t_0+s^*)|\leq |t^*-\bar t_T|+|\bar t_T-(t_0+s^*)|\leq 1+\frac{2C_{S_0}}{m_{S_0}}e^{-\delta T}.
    \end{align*}
    Absorbing $2C_{S_0}/m_{S_0}$ into $C$ and recalling $t_0=\mu_{j^*}' T/(\mu_{i^*}+\mu'_{j^*})$ gives the stated conclusion.
    \hfill $\blacksquare$

\subsection{Proof of \cref{thm:improvement}}
    Recall that we write $f_{\eh}(t)=\normh{\bm g(t)}^2$, where
    \begin{align*}
        \bm g(t)=(\bm I-e^{-(T-t)\bm G'})e^{-t\bm G} \bm \ww - e^{-(T-t) \bm G'}\bm{\Delta}.
    \end{align*}
    Unlike in the Proof of \cref{thm:localisationl2}, here $f_{\eh}:[T]\rightarrow[0,\infty)$, that is, we do not work again with a differentiable interpolator.

    For the \textbf{lower-tail} $t\in[M]$, define
    \begin{align*}
        m_{M}=\underset{t\in [M]}{\min}\normh{e^{-t \bm G}\ww}\ ,
    \end{align*}
    and note that, as $\bm G,\eh$ are both positive definite and $\ww\neq 0$ as $\mathcal I\neq \emptyset$, it must be the case that $m_M>0$. Now, using the eigendecomposition of $\eh$ and the fact that $\eh\succeq \gamma'\bm I$, for any $\bm u\in\mathbb{R}^d$:
    \begin{align*}
        \normh{e^{-(T-t)\bm G'}\bm u}^2=\sum_{j=1}^{r'}e^{-2(T-t)\mu'_j}\normh{\bm P'_j \bm u}^2\leq e^{-2(T-t)\mu_1'}\normh{\bm u}^2\ ,
    \end{align*}
    and similarly
    \begin{align*}
        \normh{(\bm I-e^{-(T-t)\bm G'})\bm u}^2\geq(1-e^{-(T-t)\mu_1'})^2\normh{\bm u}^2.
    \end{align*}
    Applying these two inequalities for $\bm u=e^{-t\bm G}\bm \ww$ and $\bm u=\bm\Delta$, and using the triangle inequality, the definition of $m_{M}$ and $T-t\geq T-M$,
    \begin{align*}
        \normh{\bm g(t)}&\geq\normh{(\bm I-e^{-(T-t)\bm G'})e^{-t\bm G}\ww}-\normh{e^{-(T-t)\bm G'}\bm\Delta}\\
        &\geq (1-e^{-\mu_1'(T-t)})\normh{e^{-t\bm G}\ww}-e^{-\mu_1'(T-t)}\normh{\bm \Delta}\\
        &\geq (1-e^{-\mu_1'(T-M)})m_{M}-e^{-\mu_1'(T-M)}\normh{\bm \Delta}.
    \end{align*}
    We can pick $T$ large enough so that
    \begin{align*}
        e^{-\mu_1'(T-M)}\leq \min\left(\frac{1}{8},\frac{m_M}{4\normh{\bm\Delta}}\right),
    \end{align*}
    for which we will have, for every $t\in[M]$
    \begin{align*}
        \normh{\bm g(t)}\geq \left(1-\frac{1}{8}\right)m_{M}-\frac{1}{4}m_M=\frac{5}{8}m_M.
    \end{align*}
    Thus, as $f_{\eh}(t)=\normh{\bm g(t)}^2$, we arrive at
    \begin{align}
        \label{eq: lower tail improvement}
        \underset{t\in [M]}{\min}f_{\eh}(t)\geq \frac{25}{64}m_M^2.
    \end{align}

    \medskip

    For the \textbf{upper-tail} $t\in \{T-M,\ldots,T\}$, we first start with the triangle inequality:
    \begin{align}
    \label{eq: normh gt}
        \normh{\bm g(t)}\geq\normh{e^{-(T-t)\bm G'}\bm \Delta}-\normh{(\bm I-e^{-(T-t)\bm G'})e^{-t\bm G}\ww}.
    \end{align}
    We bound each term on the right hand side individually. For the first term, %
    \begin{align*}
        \normh{e^{-(T-t)\bm G'}\bm \Delta}^2&=\sum_{j\in\mathcal{I}'}e^{-2(T-t)\mu'_j}\normh{\bm P'_j\bm \Delta}^2\geq e^{-2(T-t)\mu'_{j^*}}\normh{\bm P'_{j^*}\bm\Delta}^2\\
        &\geq e^{-2\mu'_{j^*}M}\normh{\bm P'_{j^*}\bm\Delta}^2,
    \end{align*}
    so, after taking the square root,
    \begin{align*}
        \normh{e^{-(T-t)\bm G'}\bm \Delta}\geq e^{-\mu'_{j^*}M}\normh{\bm P'_{j^*}\bm\Delta}.
    \end{align*}
    For the second term, denoting by $\norm{\bm H'}$ the spectral norm of $\eh$, we proceed as
    \begin{align*}
        \normh{(\bm I-e^{-(T-t)\bm G'})e^{-t\bm G}\ww}&\leq \normh{e^{-t\bm G}\ww}\leq \norm{\bm H'}^{1/2}\norm{e^{-t\bm G}\ww}\\
        &\leq \norm{\bm H'}^{1/2}e^{-\mu_1(T-M)}\norm{\ww}.
    \end{align*}
    Substituting back into \cref{eq: normh gt} we arrive at
    \begin{align*}
        \normh{\bm g(t)}\geq e^{-\mu'_{j^*}M}\normh{\bm P'_{j^*}\bm\Delta}-\norm{\eh}^{1/2}e^{-\mu_1(T-M)}\norm{\ww}.
    \end{align*}
    Now, picking $T$ large enough so that
    \begin{align*}
        \norm{\bm H'}^{1/2}e^{-\mu_1 (T-M)}\norm{\ww}\leq \frac{1}{2}e^{-\mu'_{j^*}M}\normh{\bm P'_{j^*}\bm \Delta}\ ,
    \end{align*}
    we get that
    \begin{align*}
        \normh{\bm g(t)}\geq \frac{1}{2}e^{-\mu'_{j^*}M}\normh{\bm P'_{j^*}\bm \Delta}\ ,
    \end{align*}
    and so
    \begin{align}
    \label{eq: right tail improvement}
        \underset{t\in \{T-M,\ldots,T\}}{\min}f_{\eh}(t)\geq \frac{1}{4}e^{-2\mu'_{j^*}M}\normh{\bm P'_{j^*}\bm \Delta}^2.
    \end{align}

    \medskip

    \textbf{Combining the two tails} from \cref{eq: lower tail improvement,eq: right tail improvement}, we have
    \begin{align}
        \min\{f_{\eh}(t):t\in \{0,\ldots,M\}\cup \{T-M,\ldots,T\}\}\geq\underbrace{ \min\left(\frac{25}{64}m_M^2,\frac{1}{4}e^{-2\mu'_{j^*}M}\normh{\bm P'_{j^*}\bm \Delta}^2\right)}_{:=C_M}.\label{eq:extrem_min}
    \end{align}
    We next bound $f_{\eh}$ at an arbitrary feasible allocation $t\in[T]$:
    
    Since the eigenvalues of $\bm I-e^{-(T-t)\bm G'}$ lie in $[0,1]$ and $G'$ and $H'$ have the same eigenspaces, for every $\bm v\in\mathbb{R}^d$,
    \begin{align*}
        \normh{(\bm I-e^{-(T-t)\bm G'})\bm v}\leq \normh{\bm v}.
    \end{align*}
    Moreover, since $\bm P_i\ww=0$ for $i\notin\mathcal I$,
    \begin{align*}
        \norm{e^{-t\bm G}\ww}^2=\sum_{i\in \mathcal{I}} e^{-2\mu_i t}\norm{\bm P_i\ww}^2\leq e^{-2\mu_{i^*}t}\norm{\ww}^2,
    \end{align*}
    and similarly
    \begin{align*}
        \normh{e^{-(T-t)\bm G'}\bm\Delta}^2=\sum_{j\in\mathcal I'}e^{-2\mu_j'(T-t)}\normh{\bm P_j'\bm\Delta}^2\leq e^{-2\mu'_{j^*}(T-t)}\normh{\bm \Delta}^2.
    \end{align*}
    These three bounds, together with the triangle inequality in the expression of $g(t)$, give
    \begin{align}
        \sqrt{f_{\eh}(t)}&\leq \normh{(\bm I-e^{-(T-t)\bm G'})e^{-t\bm G}\ww}+\normh{e^{-(T-t)\bm G'}\bm\Delta} \nonumber\\
        &\leq e^{-\mu_{i^*}t}\norm{\eh}^{1/2}\norm{\ww}+e^{-\mu'_{j^*}(T-t)}\normh{\bm\Delta}. \label{eq: any t upper bound}
    \end{align}
    
    Now, we can equivalently rewrite $\widehat t$ in the theorem statement as $\widehat t=\widehat \theta T+\varepsilon$ with $|\varepsilon|\leq 1/2$. Substituting this into \cref{eq: any t upper bound} and recalling that $m(\widehat\theta)=\min\{\mu_{i^*}\widehat\theta,\mu'_{j^*}(1-\widehat\theta)\}$, we have
    \begin{align*}
        \sqrt{f_{\eh}(\widehat{t})}&\leq e^{-\mu_{i^*}\widehat\theta T-\mu_{i^*}\varepsilon}\norm{\eh}^{1/2}\norm{\ww}+e^{-\mu'_{j^*}(1-\widehat{\theta})T+\mu'_{j^*}\varepsilon}\normh{\bm\Delta}\\
        &\leq e^{-m(\widehat{\theta})T}\left(\underbrace{\norm{\eh}^{1/2}\norm{\ww}e^{\mu_{i^*}/2}+\normh{\bm\Delta}e^{\mu'_{j^*}/2}}_{:=C'}\right).
    \end{align*}
    Note that $C'$ is independent of $\widehat\theta$ and $T$. Combining this bound with \cref{eq:extrem_min}, we arrive at
    \begin{align*}
        f_{\eh}(\widehat{t})\leq \frac{(C')^2}{C_M}e^{-2m(\widehat{\theta})T}\min\{f_{\eh}(t):t\in \{0,\ldots,M\}\cup \{T-M,\ldots,T\}\},
    \end{align*}
    and the inequality in the theorem statement follows by renaming $C_1:=(C')^2/C_M$.

    Finally, the $\widehat{\theta}$ that maximises the rate $m(\widehat{\theta})$ equates the two terms in the minimum defining $m(\widehat{\theta})$. This is precisely $\widehat{\theta}=\theta^*$ defined in (\ref{def theta}).
\hfill$\blacksquare$

\subsection{Proof of \cref{thm:localisation}}
Although this proof is similar to the Proof of \cref{thm:localisationl2}, with the exception of Step 5, we provide it in full below.

We split the proof into several steps. Throughout the Steps 1-6 of the proof, we will work with the interpolator of $f_{\esp}$, which we will define by overloading
\begin{align*}
    f_{\esp}:[0,T]\rightarrow[0,\infty),\quad f_{\esp}(t)=\norms{(\bm I-e^{-(T-t)\bm G'})e^{-t \bm G}\ww - e^{-(T-t)\bm G'}\bm{\bm{\Delta}}}^2.
\end{align*}
We will revert back to the original discrete $f_{\esp}:[T]\rightarrow[0,\infty)$ at the end, in Step 7.

\noindent \textbf{Step 1: Decomposing $f_{\esp}'$:} Define 
\begin{align*}
    \bm g(t):=(\bm I-e^{-(T-t)\bm G'})e^{-t \bm G}\ww - e^{-(T-t)\bm G'}\bm{\Delta}\,.
\end{align*}
Using the spectral transformations of $\bm G$ and $\bm G'$, we have
\begin{align*}
    \bm g(t)&=\sum_{i=1}^r e^{-\mu_i t} \bm P_i\ww-\sum_{i=1}^r\sum_{j=1}^{r'}e^{-\mu_i t-\mu'_j(T-t)}\bm P'_j \bm P_i\ww-\sum_{j=1}^{r'}e^{-\mu'_j(T-t)}\bm P'_j\bm{\Delta}\,,\\
    \bm g'(t)&=-\sum_{i=1}^r\mu_ie^{-\mu_i t} \bm P_i\ww+\sum_{i=1}^r\sum_{j=1}^{r'}(\mu_i-\mu'_j)e^{-\mu_i t-\mu'_j(T-t)}\bm P'_j \bm P_i\ww-\sum_{j=1}^{r'}\mu'_je^{-\mu'_j(T-t)}\bm P'_j\bm{\Delta}\,.
\end{align*}
Further, we have $f_{\esp}'(t)=2 \bm g(t)^\top \esp \bm g'(t)$. 

For ease of presentation, we use the following notations:
\begin{align*}
    c^*:=\frac{\mu'_{j^*}}{\mu_{i^*}+\mu'_{j^*}}, \quad t_0:=c^*T, \quad m^*:=\frac{\mu_{i^*}\mu'_{j^*}}{\mu_{i^*}+\mu'_{j^*}},\quad\kappa^*:=e^{-2m^*T}\,.
\end{align*}

Using the transformation $t=t_0+s$ with $s\in [-t_0,T-t_0]$, we can decompose
\begin{align*}
    \bm g(t_0+s)&=e^{-\mu_{i^*}(t_0+s)}\bm P_{i^*}\ww - e^{-{\mu'_{j^*}}(T-(t_0+s))} \bm P'_{j^*}\bm{\Delta}+\sum_{i\neq i^*,i=1}^{r}e^{-\mu_i (t_0+s)} \bm P_i\ww\\
    &-\sum_{i=1}^r\sum_{j=1}^{r'}e^{-\mu_i (t_0+s)-\mu'_j(T-(t_0+s))}\bm P'_j \bm P_i\ww
     - {\sum_{j\neq j^*,j=1}^{r'}e^{-\mu'_j(T-(t_0+s))}\bm P'_j\bm{\Delta}} \quad \\ %
    &:=\underbrace{e^{-\mu_{i^*}(t_0+s)}\bm P_{i^*}\ww-e^{-{\mu'_{j^*}}(T-(t_0+s))} \bm P'_{j^*}\bm{\Delta}}_{\bm g_{\operatorname{lead}}}+\bm g_{\operatorname{rest}}(t_0+s),\\
    \bm g'(t_0+s)&=\underbrace{-\mu_{i^*}e^{-\mu_{i^*}(t_0+s)}\bm P_{i^*}\ww-\mu'_{j^*}e^{-\mu'_{j^*}(T-(t_0+s))} \bm P'_{j^*}\bm{\Delta}}_{\bm g_{\operatorname{lead}}'}+\bm g'_{\operatorname{rest}}(t_0+s)\,,
\end{align*}
and further, taking the inner product between these two, we have
\begin{align*}
    \frac{f_{\esp}'(t_0+s)}{2}&=\gl(t_0+s)^\top \esp \glp(t_0+s)\\
    &+\underbrace{(\gr^\top \esp \glp+\gl^\top \esp\grp+\gr^\top \esp\grp)(t_0}_{:=\kappa^*\cdot \Xi}+s)\,,
\end{align*}
where we have introduced, for notation simplicity, the function $\Xi:[-t_0,T-t_0]\rightarrow\mathbb{R}$.%

Now, note that $\sqrt{\kappa^*}=e^{-m^*T}=e^{-\mu_{i^*}t_0}=e^{-\mu'_{j^*}(T-t_0)}$, so that
\begin{align*}
    \Gamma(s)&:=\frac{\gl(t_0+s)^\top \esp \glp(t_0+s)}{\kappa^*}\\
    &=\mu'_{j^*}\norms{\bm P'_{j^*}\bm{\Delta}}^2e^{2\mu'_{j^*} s}\!-\!\mu_{i^*}\norms{\bm P_{i^*}\ww}^2e^{-2\mu_{i^*} s}\!-\!(\mu'_{j^*}\!-\!\mu_{i^*}) \ww^\top \bm P_{i^*}\esp \bm P'_{j^*}\bm{\Delta} e^{(\mu'_{j^*}-\mu_{i^*})s},
\end{align*}
and further we have decomposed $f'$ into (as we show below) a quadratic $\Gamma$ in the exponential space plus a perturbation term $\Xi$:
\begin{align}
\label{f' decomp}
    \frac{f_{\esp}'(t_0+s)}{2\kappa^*}=\Gamma(s)+\Xi(s).
\end{align}

We next compute the single zero of $\Gamma$ and show that around this zero the function $\Gamma$ is increasing: 
    
\bigskip

    \noindent\textbf{Step 2: Analysing $\Gamma(s)$:} Recall the notations
    \begin{align}
    \label{abd}
        a^*=\norms{\bm P_{i^*}\ww}^2,\quad b^*:=\norms{\bm P'_{j^*}\bm{\Delta}}^2,\quad d^*:=\ww^\top \bm P_{i^*}\esp \bm P'_{j^*}\bm{\Delta}.
    \end{align}
    With $q:=e^{(\mu_{i^*}+\mu'_{j^*})s}>0$, we have %
    \begin{align*}
        e^{2\mu_{i^*} s}\Gamma(s)=\mu'_{j^*}b^*q^2-(\mu'_{j^*}-\mu_{i^*})d^* q -\mu_{i^*} a^*.
    \end{align*}
    This is a quadratic in $q>0$. The product of the roots is negative, so the quadratic has exactly one positive root. Taking its $\log$ and dividing by $\mu_{i^*}+\mu'_{j^*}$\ we arrive at the unique zero of $\Gamma$:
    \begin{align*}
        s^*=\frac{1}{\mu_{i^*}+\mu'_{j^*}}\ln\left(\frac{(\mu'_{j^*}-\mu_{i^*})d^*+\sqrt{(\mu'_{j^*}-\mu_{i^*})^2(d^*)^2+4\mu_{i^*}\mu'_{j^*}a^*b^*}}{2\mu'_{j^*} b^*}\right).
    \end{align*}
    Next, using the fact that $\Gamma(s^*)=0 \iff (\mu'_{j^*}-\mu_{i^*})d^* e^{(\mu'_{j^*}-\mu_{i^*})s^*}=\mu'_{j^*}b^*e^{2\mu'_{j^*}s^*}-\mu_{i^*}a^*e^{-2\mu_{i^*}s^*}$, we have%
    \begin{align*}
        \Gamma'(s^*)&=2(\mu'_{j^*})^2b^*e^{2\mu'_{j^*}s^*}+2(\mu_{i^*})^2a^*e^{-2\mu_{i^*}s^*}-(\mu'_{j^*}-\mu_{i^*})^2d^*e^{(\mu'_{j^*}-\mu_{i^*})s^*}\\
        &=2(\mu'_{j^*})^2b^*e^{2\mu'_{j^*}s^*}+2(\mu_{i^*})^2a^*e^{-2\mu_{i^*}s^*}-(\mu'_{j^*}-\mu_{i^*})(\mu'_{j^*}b^*e^{2\mu'_{j^*}s^*}-\mu_{i^*}a^*e^{-2\mu_{i^*}s^*})\\
        &=(\mu_{i^*}+\mu'_{j^*})(\mu'_{j^*}b^*e^{2\mu'_{j^*}s^*}+\mu_{i^*}a^*e^{-2\mu_{i^*}s^*})>0\,.
    \end{align*}
    So, there exists $S_0>0$ (independent of $T$) such that $m_{S_0}:=\min_{|s-s^*|\leq S_0}\Gamma'(s)>0$. In this interval $[s^*-S_0,s^*+S_0]$ the function $\Gamma$ is thus increasing. It is negative on $[s^*-S_0,s^*)$ and positive on $(s^*,s^*+S_0]$. Moreover, one can pick $T$ large enough to make sure $[t_0+s^*-S_0,t_0+s^*+S_0]\subseteq [0,T]$.

    Next, we show that the additive perturbation term $\Xi$ is small on any compact interval around $s^*$:

    \bigskip

    \noindent\textbf{Step 3: Uniform bound for the perturbation term $\Xi(s)$ in any compact $[s^*-S,s^*+S]$:} We show that for any $S>0$ (so, in particular, for the previous choice of $S_0$), there exist $C_S>0$ that may change from line to line -- but stay independent of $T$ and just absorb other constants -- and $\delta>0$ (also independent of $T$) such that for $T$ large enough
    \begin{align*}
        \underset{|s-s^*|\leq S}{\sup}(|\Xi(s)|+|\Xi'(s)|)\leq C_Se^{-\delta T}.
    \end{align*}
    Recalling the definition of $\Xi$ from \cref{f' decomp} and using Cauchy-Schwarz inequality, %
    \begin{align}
    \label{|P| ridge}
       |\Xi|&\leq \frac{1}{\kappa^*}\left( \norms{\bm g_{\operatorname{rest}}}\norms{\bm g'_{\operatorname{lead}}}+\norms{\bm g_{\operatorname{lead}}}\norms{\bm g'_{\operatorname{rest}}}+\norms{\bm g_{\operatorname{rest}}}\norms{\bm g'_{\operatorname{rest}}}\right). %
    \end{align}
    We next bound each $\esp$-seminorm. For $s$ with $|s-s^*|\leq S$, as $|s|\leq |s^*|+S:=M_S$, we have
    \begin{align*}
        \left|e^{\pm\mu_i s}\right|=e^{\pm \mu_i s}\leq e^{\mu_i|s|}\leq e^{\mu_i M_S}\leq e^{\mu_r M_S}\leq e^{(\mu_{r}+\mu'_{r'})M_S}=:C_S,
    \end{align*}
    and similarly $|e^{\pm \mu'_j s}|\leq C_S$, $|e^{\pm(\mu'_j-\mu_i)s}|\leq C_S$.

    Recall the notation $\bm g_{\operatorname{lead}}(t_0+s)=e^{-\mu_{i^*}(t_0+s)}\bm P_{i^*}\ww-e^{-\mu'_{j^*}(T-(t_0+s))} \bm P'_{j^*}\bm{\Delta}$ and $t_0=c^*T=\mu'_{j^*}T/(\mu_{i^*}+\mu'_{j^*})$ and the notations introduced in \cref{abd}. Then, by a triangle inequality,%
    \begin{align}
    \label{g lead}
            \norms{\bm g_{\operatorname{lead}}(t_0+s)}&\leq e^{-\mu_{i^*} t_0}e^{-\mu_{i^*} s}\norms{\bm P_{i^*}\ww}+e^{-\mu'_{j^*}(T-t_0)}{e^{\mu'_{j^*} s}\norms{\bm P'_{j^*}\bm{\Delta}}}\nonumber\\
            &=e^{-m^* T}e^{-\mu_{i^*} s}\norms{\bm P_{i^*}\ww}+e^{-m^* T}e^{\mu'_{j^*} s}\norms{\bm P'_{j^*}\bm{\Delta}} \leq C_S e^{-m^* T}.
    \end{align}
     A similar computation (by letting $C_S$ absorb $\mu_{i^*}$ and $\mu'_{j^*}$) gives $\norms{\bm g'_{\operatorname{lead}}(t_0+s)}\leq C_S e^{-m^* T}$.

    By the definition of $\mu_{i^*}$ and $\mu'_{j^*}$, $\delta_1:=\underset{i\in \mathcal{I}\setminus\{i^*\}}{\min}(c^*(\mu_i-\mu_{i^*}))>0$ (taken to be $+\infty$ if $\mathcal{I}=\{i^*\}$), $\delta_2:=\underset{j\in \mathcal I'\setminus\{j^*\}}{\min}((1-c^*)(\mu'_j-\mu'_{j^*}))>0$ (taken to be $+\infty$ if $\mathcal{I}'=\{j^*\}$). Additionally, as $\bm H'$ is positive definite, $(1-c^*)\mu'_1>0$. We take $\delta:=\min(\delta_1,\delta_2,(1-c^*)\mu'_1)>0$.

     Next, we deal with bounding $\bm g_{\operatorname{rest}}$ and its derivative. However, for the derivative, we can simply absorb the eigenvalues into the constants, the analysis is otherwise identical. There are three types of contribution in it:

     \begin{itemize}
         \item From the first summation, terms with $i\notin \mathcal{I}$ have zero seminorm. For $i\neq i^*$, $i\in \mathcal{I}$, using the fact that $\mu_{i^*}c^*=m^*$, these terms can be bounded as
         \begin{align*}
             \norms{e^{-\mu_i(t_0+s)} \bm P_i\ww}&=e^{-\mu_i t_0}e^{-\mu_i s}\norms{\bm P_i\ww }\leq e^{-\mu_ic^*T}C_S\norms{P_i\ww}\\
             &=e^{-(\mu_i-\mu_{i^*})c^*T}e^{-\mu_{i^*} c^* T}C_S\norms{\bm P_i\ww}\\
             &\leq C_S e^{-\delta T}e^{-m^* T}\norms{\bm P_i\widehat{\bm w}}=C_Se^{-(m^*+\delta)T}.
         \end{align*}
         where the constant (independent of $T$) $C_S$ absorbed $\norms{\bm P_i\ww}$.
         \item Similarly, from the third summation, the terms with $j\neq j^*$ can be bounded as
         \begin{align*}
             \norms{e^{-\mu'_j(T-(t_0+s))}\bm P'_j\bm{\Delta}}\leq C_Se^{-(m^*+\delta)T}. 
         \end{align*}
         \item For the second summation over both $i\in \{1,\ldots,r\}$ and $j\in \{1,\ldots,r'\}$, the contribution of each pair $(i,j)$ is:
         \begin{align*}
             \norms{e^{-\mu_i(t_0+s)-\mu'_j(T-(t_0+s))}\bm P'_j \bm P_i\ww}\leq C_Se^{-(\mu_ic^*+\mu'_j(1-c^*))T}.
         \end{align*}
         Now, note that if $i\not\in\mathcal I$, the contribution of $(i,j)$ is $0$: as $\boldsymbol P'_j$ commutes with $\esp$, we have $\esp\bm P'_j\boldsymbol P_i\widehat{\boldsymbol w}=\boldsymbol P'_j\esp\boldsymbol P_i\widehat{\boldsymbol w}=\bm 0$. Moreover, for $i\in \mathcal{I}$ we have
         \begin{align*}
             \mu_ic^*+\mu'_j(1-c^*)&\geq \mu_{i^*}c^*+\mu'_1(1-c^*)=m^*+\mu'_1(1-c^*)\geq m^*+\delta\,,
         \end{align*}
         so that the contribution for each pair $(i,j)$ is again at most $C_Se^{-(\delta+m^*)T}$.
     \end{itemize}
     Putting everything together and absorbing the sum of all constants into a single constant $C_S$ (which depends on $r$, $r'$), we get that for any $s\in [s^*-S,s^*+S]$,
     \begin{align}
     \label{g rest}
         \norms{\bm g_{\operatorname{rest}}(t_0+s)}\leq C_S e^{-(m^*+\delta)T}\quad\text{and}\quad \norms{\bm g'_{\operatorname{rest}}(t_0+s)}\leq C_S e^{-(m^*+\delta)T},
     \end{align}
     for some appropriate (independent of $T$, but containing $r$ and $r'$) constant $C_S$.

     Using \cref{|P| ridge}, we get that
     \begin{align*}
         |\Xi(s)|\leq \frac{C_S}{\kappa^*}(e^{-(2m^*+\delta)T}+e^{-(2m^*+\delta)T}+e^{-2(m^*+\delta)T})\leq C_Se^{-\delta T}. 
     \end{align*}

     The bound for $\Xi'(s)$ is almost identical, with the only difference being that
     \begin{align*}
         \kappa^* \Xi'=2(\bm g'_{\operatorname{rest}})^\top\esp \bm g'_{\operatorname{lead}}+ \bm g_{\operatorname{rest}}^\top\esp \bm g''_{\operatorname{lead}}+(g''_{\operatorname{rest}})^\top\esp g_{\operatorname{lead}}+\norms{\bm g'_{\operatorname{rest}}}^2+\bm g_{\operatorname{rest}}^\top \esp g''_{\operatorname{rest}},
     \end{align*}
     and now the bounds for the second derivatives are (up to constants independent of $T$) identical.

     Given this bound on the perturbation term $\Xi$ on any compact interval around $s^*$, we show that it cannot perturb the zero of $\Gamma(s)$ by too much:

    \bigskip
     \noindent\textbf{Step 4: Existence, uniqueness and localisation of the zero of $\Gamma(s)+\Xi(s)$ inside $[s^*-S_0,s^*+S_0]$:} First, we show that $\Gamma(s)+\Xi(s)$ has a unique zero in $(s^*-S_0,s^*+S_0)$ for $T$ large enough. Start by defining
     \begin{align*}
         \eta_{S_0}:=\min\{-\Gamma(s^*-S_0),\Gamma(s^*+S_0)\}
     \end{align*}
     and note that $\eta_{S_0}>0$, as the derivative in $[s^*-S_0,s^*+S_0]$ is positive by the choice of $S_0$ and $\Gamma(s^*)=0$. Moreover, pick $T_0$ large enough such that, for all $T\geq T_0$,
     \begin{align*}
         C_{S_0}e^{-\delta T}\leq \min\{\eta_{S_0}/2,m_{S_0}/2\},
     \end{align*}
     where we recall that $m_{S_0}:=\min_{|s-s^*|\leq S_0}\Gamma'(s)>0$ and $C_{S_0}$ is the constant from the previous Step 3 when instantiating $S:=S_0$. This is possible as $S_0$ is independent of $T$. Then, for all $s\in [s^*-S_0,s^*+S_0]$, $|\Xi(s)|\leq \eta_{S_0}/2$ and $|\Xi'(s)|\leq m_{S_0}/2$. This implies that on that interval
     \begin{align}
     \label{Q'+P' esp}
         \Gamma'(s)+\Xi'(s)\geq m_{S_0}-|\Xi'(s)|\geq m_{S_0}/2>0,
     \end{align}
     so if there is a zero in this interval, it is unique. And there is indeed a zero: at the endpoints
     \begin{align*}
         &\Gamma(s^*-S_0)+\Xi(s^*-S_0)\leq -\eta_{S_0}+\eta_{S_0}/2=-\eta_{S_0}/2<0 \quad \text{and}\quad \\ &\Gamma(s^*+S_0)+\Xi(s^* + S_0)\geq \eta_{S_0}-\eta_{S_0}/2=\eta_{S_0}/2>0.
     \end{align*}
    and the existence follows by the Intermediate Value Theorem. Let $z_T$ be this unique zero. Since $f'(s+t_0)=2\kappa^* (\Gamma(s)+\Xi(s))$ is negative at the left end of the interval, $0$ at $z_T$, positive at the right end, it follows that $z_T+t_0$ is the minimum of $f$ in this interval.
    
    Further, note that $z_T$ satisfies $(\Gamma+\Xi)(z_T)=0$  and recall that $\Gamma(s^*)=0$. Then, by the Mean Value Theorem on the interval $I\subseteq [s^*-S_0,s^*+S_0]$ with endpoints $z_T$ and $s^*$,
    \begin{align*}
        |\Xi(s^*)|&=|(\Xi+\Gamma)(s^*)-(\Xi+\Gamma)(z_T)|\geq |s^*-z_T|\underset{\xi\in \mathcal{I}}{\inf}|\Gamma'(\xi)+\Xi'(\xi)|\\
        &\geq |s^*-z_T|\underset{\xi\in [s^*-S_0,s^*+S_0]}{\inf}|\Gamma'(\xi)+\Xi'(\xi)|\geq |s^*-z_T|m_{S_0}/2,
    \end{align*}
    where the last inequality follows from \cref{Q'+P' esp}. Next, recall that $|\Xi(s^*)|\leq C_{S_0}e^{-\delta T}$, so that
    \begin{align*}
        |z_T-s^*|\leq \frac{2C_{S_0}}{m_{S_0}}e^{-\delta T}.
    \end{align*}
    Defining $\bar{t}_T:=t_0+z_T$, we arrive at
    \begin{align*}
        |\bar{t}_T-(t_0+s^*)|\leq \frac{2C_{S_0}}{m_{S_0}}e^{-\delta T}.
    \end{align*}
    It remains to locate the discrete global minimisers of $f_{\esp}$\ . We do this as follows: Step 5 proves that these are within $[t_0+S_-,t_0+S_+]$, Step 6 establishes the sign of $f'_{\esp}$ on this central interval, and finally Step 7 shows that these global minimisers lie at the integers adjacent to $\bar t_T$.

    \bigskip
    \noindent\textbf{Step 5: Proving the discrete global minimum lies inside $t_0+[S_-,S_+]$ for some $S_-<0$, $S_-<s^*<S_+$:} 

    \textbf{Upper-bounding $f_{\esp}(t_0+s)$ around $s^*$:} Recall that $f_{\esp}(t_0+s)=\norms{\bm g(t_0+s)}^2$. Define
    \begin{align*}
        \hat t_T\in\arg\min_{q\in [T]}|q-(t_0+s^*)|,\quad \hat s_T:=\hat t_T-t_0.
    \end{align*}
    For a sufficiently large $T$, we have $|\hat{s}_T-s^*|\leq 1/2$.
    
    Now, we can apply the bounds derived in Step 3. To be more precise, consider $S=1$, then, according to the triangle inequality, \cref{g lead} and \cref{g rest}:
    \begin{align*}
        \sup_{|s-s^*|\leq 1}f_{\esp}(t_0+s)&=\sup_{|s-s^*|\leq 1}{\norms{\bm g(t_0+s)}^2}\\ &\leq \sup_{|s-s^*|\leq 1} (\norms{\gl(t_0+s)}+\norms{\gr(t_0+s)})^2\\
        &\leq C_1^2(e^{-m^* T}+e^{-(m^*+{\delta})T})^2
        \leq 2C_1^2 e^{-2m^* T}=2C_1^2 \kappa^*.
    \end{align*}
    Importantly, the constant does not depend on $T$. Since $s:=\hat s_T$ satisfies $|s-s^*|\leq 1$, this inequality gives 
    \begin{align}
        \label{step 5 discrete esp}
        f_{\esp}(\hat t_T)\leq 2C_1^2\kappa^*.
    \end{align}

    \textbf{Left-tail lower bound:} Note that if $t=t_0+s\geq 0$, then $s\geq -t_0$. We argue that for a certain $S_-<s^*$ (independent of $T$) and all $T$ large enough (depending on $S_-$) we have
    \begin{align*}
        f_{\esp}(t_0+s)\geq \frac{81}{1024}e^{-2\mu_{i^*} S_-}\norms{\bm P_{i^*}\ww}^2\kappa^*
    \end{align*}
    for every integer $t_0+s\in[T]$ with $s\leq S_-$. 
    
    By the triangle inequality,
    \begin{align}
    \label{lower norms gt}
        \norms{\bm g(t)}&\geq \norms{(\bm I-e^{-(T-t)\bm G'})e^{-t \bm G}\ww}-\norms{e^{-(T-t)\bm G'}\bm{\Delta}}.
    \end{align}
    We now bound each of the norms.
    
    Since $\bm H'=\esp+\gamma' \bm I$ has eigenvalue/projector pairs $(\lambda'_j,\bm P'_j)_{j=1}^{r'}$\ , the eigenvalue/projector pairs for $\esp$ are  $(\lambda'_j-\gamma',\bm P'_j)_{j=1}^{r'}$. Now, for an arbitrary vector $\bm v$,
    \begin{align}
    \label{sigma v}
        \norms{e^{-(T-t)\bm G'}\bm v}^2&=(e^{-(T-t)\bm G'}\bm v)^\top \esp (e^{-(T-t)\bm G'}\bm v) \nonumber\\
        &=\left(\sum_{j=1}^{r'}e^{-\mu'_j(T-t)}\bm P'_j \bm v\right)^\top\left(\sum_{j=1}^{r'}(\lambda'_j-\gamma') \bm P'_j\right)\left(\sum_{j=1}^{r'}e^{-\mu'_j(T-t)}\bm P'_j \bm v\right) \nonumber\\
        &=\sum_{j=1}^{r'}(\lambda'_j-\gamma')e^{-2\mu'_j(T-t)}\bm v^\top \bm P'_j \bm v
        \\&\leq e^{-2\mu_1'(T-t)}\sum_{j=1}^{r'}(\lambda'_j-\gamma')\bm v^\top \bm P'_j \bm v
        =e^{-2\mu_1'(T-t)}\norms{\bm v}^2.\nonumber
    \end{align}
    Now, applying this to $\bm v=e^{-t \bm G}\ww$ and using the triangle inequality,
    \begin{align*}
        \norms{(\bm I-e^{-(T-t)\bm G'})e^{-t \bm G}\ww}&\geq \norms{e^{-t \bm G}\ww}-\norms{e^{-(T-t)\bm G'}e^{-t \bm G}\ww}\\
        &\geq (1-e^{-(T-t)\mu_1'})\norms{e^{-t \bm G}\ww}.
    \end{align*}
    Next, consider $S_-<s^*$ to be fixed later and recall that we translated $t=t_0+s$. For $s\leq S_-$, $T-(t_0+s)\geq (1-c^*)T-S_-\rightarrow\infty$ as $T\rightarrow\infty$. So, we can pick $T$ large enough (and dependent on $S_-$) such that
    \begin{align}
    \label{i-e}
        \norms{(I-e^{-(T-t)\bm G'})e^{-t \bm G}\ww}&\geq\frac{3}{4}\norms{e^{-t \bm G}\ww}.
    \end{align}
    
    For the other norm in \cref{lower norms gt}, continuing from \cref{sigma v} for $v=\bm{\Delta}$,
    \begin{align}
    \label{norm eth delta}
        \norms{e^{-(T-t)\bm G'}\bm{\Delta}}^2&= \sum_{j=1}^{r'}(\lambda'_j-\gamma')e^{-2\mu'_j(T-t)}\bm{\Delta}^\top \bm P'_j \bm\Delta=\sum_{j=1}^{r'}e^{-2\mu'_j (T-t)}\norms{\bm P'_j\bm{\Delta}}^2\nonumber\\
        &=\sum_{j\in \mathcal I'}e^{-2\mu'_j (T-t)}\norms{\bm P'_j\bm{\Delta}}^2\\
        &\leq e^{-2\mu'_{j^*}(T-t)}\sum_{j\in \mathcal I'}\norms{\bm P'_j\bm{\Delta}}^2=e^{-2\mu'_{j^*}(T-t)}\norms{\bm{\Delta}}^2,\nonumber
    \end{align}
    from which $\smallnorm{e^{-(T-t)\bm G'}\bm{\Delta}}_{\esp}\leq e^{-\mu'_{j^*}(T-t)}\smallnorm{\bm{\Delta}}_{\esp}$. Substituting this and taking \cref{i-e} back into \cref{lower norms gt}, and using \cref{norm eth delta}, we arrive at
    \begin{align}
    \label{lb gt}
        \norms{\bm g(t)}&\geq \frac{3}{4}\norms{e^{-t \bm G}\ww}-\norms{e^{-(T-t)\bm G'}\bm{\Delta}} \nonumber\\
        &\geq \frac{3}{4}\norms{e^{-t \bm G}\ww}-e^{-\mu'_{j^*}(T-t)}\norms{\bm{\Delta}}
    \end{align}
    Now, we claim that when $K$ is defined as in \hyperref[additionalnotation]{Appendix~\ref*{additionalnotation}} (depending on $\bm H$, $\ww$, $\esp$ but not $T$), for all $t\in \{K,\ldots,T\}$,
    \begin{align}
    \label{lb eth}
        \norms{e^{-t \bm G}\ww}\geq\frac{3}{4}e^{-\mu_{i^*}t}\norms{\bm P_{i^*}\ww}.
    \end{align}
    If $\mathcal{I}=\{i^*\}$, this clearly holds with $K=0$. Otherwise, defining $\xi:=\min_{i\in \mathcal{I}\setminus\{i^*\}}(\mu_i-\mu_{i^*})>0$, by the triangle inequality,
    \begin{align*}
        \norms{e^{-t \bm G}\ww}&=\norms{\sum_{i=1}^{r}e^{-\mu_i t} \bm P_i\ww}\geq e^{-\mu_{i^*} t}\norms{\bm P_{i^*}\ww}-\sum_{i\neq i^*,i=1}^r e^{-\mu_i t}\norms{P_i\ww}\\
        &=e^{-\mu_{i^*} t}\norms{\bm P_{i^*}\ww}-\sum_{i\in \mathcal{I}\setminus\{i^*\}}^r e^{-\mu_i t}\norms{P_i\ww}\\
        &=e^{-\mu_{i^*} t}\left(\norms{\bm P_{i^*}\ww}-\sum_{i\in \mathcal{I}\setminus\{i^*\}}e^{-(\mu_i-\mu_{i^*})t}\norms{P_i\ww}\right)\\
        &\geq e^{-\mu_{i^*} t}\left(\norms{\bm P_{i^*}\ww}-e^{-\xi t}\sum_{i\in \mathcal{I}\setminus\{i^*\}}\norms{P_i\ww}\right).
    \end{align*}
    By the definition of $K$ in \hyperref[additionalnotation]{Appendix~\ref*{additionalnotation}}, for any $t\in \{K,\ldots,T\}$, we have $e^{-\xi t}\sum_{i\in \mathcal{I}\setminus\{i^*\}}\norms{\bm P_i\ww}\leq \norms{\bm P_{i^*}\ww}/4$, and then \cref{lb eth} holds. We next consider two different regimes:
    \begin{itemize}
        \item $t\in[K]$: We use the assumption that
        \begin{align*}
            \underset{t\in [K]}{\min}\norms{e^{-t \bm G}\ww}:=m_{K}>0.
        \end{align*}
        We consider $T$ large enough so that $e^{-\mu'_{j^*}(T-K)}\norms{\bm{\Delta}}\leq m_{K}/8$. This holds, for example, for
        \begin{align*}
            T\geq K+\frac{1}{\mu'_{j^*}}\ln\left(\frac{8\norms{\bm\Delta}}{m_{K}}\right).
        \end{align*}
        Plugging this back into \cref{lb gt}, we have
        \begin{align*}
            \norms{\bm g(t)}\geq\frac{3}{4}m_{K}-\frac{1}{8}m_{K}=\frac{5}{8}m_{K}.
        \end{align*}
        Recalling that $f_{\esp}(t)=\norms{\bm g(t)}^2$, we arrive at
        \begin{align}
        \label{left 0 tau}
            f_{\esp}(t)\geq \frac{25}{64}m_{K}^2\quad\text{for all } t\in [K],
        \end{align}
        or, equivalently for integers $t=t_0+s$ with $s\in [-t_0,K-t_0]$. As $\kappa^*=e^{-2m^* T}\rightarrow 0$ as $T\rightarrow\infty$, by choosing $T$ large enough we can make
        \begin{align*}
        f_{\esp}(t_0+s)\geq \frac{25m_{K}^2}{64}\geq \frac{9}{32}e^{-2\mu_{i^*} S_-}\norms{\bm P_{i^*}\ww}^2\kappa^* \text{ for all } s\in[K]-t_0
    \end{align*}
    for any $T$-independent $S_-$.
    \item $t$ is an integer with $K\leq t\leq t_0+S_-$: We can now apply \cref{lb eth} to \cref{lb gt}:
    \begin{align*}
        \norms{\bm g(t)}\geq \frac{9}{16}e^{-\mu_{i^*} t}\norms{\bm P_{i^*}\ww}-e^{-\mu'_{j^*}(T-t)}\norms{\bm{\Delta}}.
    \end{align*}
    Setting $t=t_0+s$, we have
    \begin{align*}
        \norms{\bm g(t_0+s)}&\geq e^{-\mu_{i^*}t_0}\left(\frac{9}{16}e^{-\mu_{i^*} s}\norms{\bm P_{i^*}\ww}-e^{-\mu'_{j^*}(T-t_0-s)+\mu_{i^*}t_0}\norms{\bm{\Delta}}\right)\\
        &=e^{-m^* T}\left(\frac{9}{16}e^{-\mu_{i^*} s}\norms{\bm P_{i^*}\ww}-e^{\mu'_{j^*} s}\norms{\bm{\Delta}}\right).
    \end{align*}
    We can adjust $S_-<0$ and $S_-<s^*$ independent of $T$ such that for all $s\leq S_-<0$,
    \begin{align*}
        e^{\mu'_{j^*} s}\norms{\bm{\Delta}}\leq \frac{9}{32}e^{-\mu_{i^*} s}\norms{\bm P_{i^*}\ww},
    \end{align*}
    which gives
    \begin{align*}
        \norms{\bm g(t_0+s)}\geq \frac{9}{32}e^{-m^* T}e^{-\mu_{i^*} s}\norms{\bm P_{i^*}\ww}\geq \frac{9}{32}e^{-m^* T}e^{-\mu_{i^*} S_-}\norms{\bm P_{i^*}\ww}.
    \end{align*}
    Squaring and using $\kappa^*=e^{-2m^* T}$, we arrive again at
    \begin{align*}
        f_{\esp}(t_0+s)\geq \frac{81}{1024}e^{-2\mu_{i^*} S_-}\norms{\bm P_{i^*}\ww}^2\kappa^*
    \end{align*}
    for every integer $t_0+s\in[T]$ with $t_0+s\geq K$ and $s\leq S_-$. 
    \end{itemize}
    If needed, we can increase $T$ to ensure that $K\leq t_0+S_-$, so that the two cases analysed cover the full discrete left tail.

    \bigskip
    \textbf{Right-tail lower-bound:} Note that if $t=t_0+s\leq T$, then $s\leq T-t_0$. We argue that for a certain $S_+>s^*$ independent of $T$ (and for $T$ large enough depending on $S_+$), we have 
    \begin{align*}
        f_{\esp}(t_0+s)\geq \frac{9}{16}\norms{\bm P'_{j^*}\bm{\Delta}}^2e^{2\mu'_{j^*} S_+}\kappa^*
    \end{align*}
    for every integer $t_0+s\in[T]$ and $s\geq S_+$. 
    
    For this, by the triangle inequality and the definition of the spectral norm,
    \begin{align}
    \label{gt upper}
        \norms{\bm g(t)}&\geq  \norms{e^{-(T-t)\bm G'}\bm{\Delta}}-\norms{(\bm I-e^{-(T-t)\bm G'})e^{-t \bm G}\ww}.
    \end{align}
    We bound each of the two norms. 
    
    For the first norm on the right-hand side, continuing from \cref{norm eth delta},
    \begin{align*}
        \norms{e^{-(T-t)\bm G'}\bm{\Delta}}^2
        &=\sum_{j\in \mathcal{I}'}e^{-2\mu'_j (T-t)}\norms{\bm P'_j\bm{\Delta}}^2\geq e^{-2\mu'_{j^*}(T-t)}\norms{\bm P'_{j^*}\bm{\Delta}}^2,
    \end{align*}
    so that
    \begin{align}
        \label{gt upper first}
        \smallnorm{e^{-(T-t)\bm G'}\bm{\Delta}}_{\esp}\geq e^{-\mu'_{j^*}(T-t)}\smallnorm{\bm P'_{j^*}\bm{\Delta}}_{\esp}.
    \end{align}
    For the second norm, as $\bm I=\sum_{j=1}^{r'} \bm P'_j$, proceeding analogously as we have arrived at \cref{sigma v},
    \begin{align}
    \label{upper i-e}
        \norms{(\bm I-e^{-(T-t)\bm G'})e^{-t \bm G}\ww}^2&=\sum_{j=1}^{r'}(\lambda'_j-\gamma')(1-e^{-\mu'_j(T-t)})^2 (e^{-t \bm G}\ww)^\top \bm P'_j e^{-t \bm G}\ww \nonumber\\
        &\leq \sum_{j=1}^{r'}(\lambda'_j-\gamma') (e^{-t \bm G}\ww)^\top \bm P'_j e^{-t \bm G}\ww \nonumber\\
        &=\norms{e^{-t \bm G}\ww}^2.
    \end{align}
    Further, taking the square root, using the eigendecomposition of $\bm G$ and the triangle inequality,
    \begin{align*}
        \norms{(\bm I-e^{-(T-t)\bm G'})e^{-t \bm G}\ww}&\leq \norms{e^{-t \bm G}\ww}=\norms{\sum_{i\in \mathcal{I}}e^{-t\mu_i} \bm P_i\ww}\\
        &\leq \sum_{i\in \mathcal{I}} e^{-t\mu_i}\norms{\bm P_i\ww}\leq e^{-\mu_{i^*}t}\sum_{i\in \mathcal{I}}\norms{P_i\ww}.
    \end{align*}
    Substituting this together with \cref{gt upper first} back into \cref{gt upper}, we arrive at
    \begin{align*}
        \norms{\bm g(t)}\geq e^{-\mu'_{j^*}(T-t)}\smallnorm{\bm P'_{j^*}\bm{\Delta}}_{\esp}-e^{-\mu_{i^*}t}\sum_{i\in \mathcal{I}}\norms{P_i\ww}.
    \end{align*}
    With the substitution $t=t_0+s$, we get
    \begin{align*}
        \norms{\bm g(t)}\geq e^{-\mu'_{j^*}(T-t_0-s)}\left(\norms{\bm P'_{j^*}\bm{\Delta}}-e^{-(\mu_{i^*}+\mu'_{j^*})s}\sum_{i\in \mathcal{I}}\norms{\bm P_i\ww}\right).
    \end{align*}
    Now, we can choose $S_+>\max(0,s^*)$ large enough so that for all $s\geq S_+$,
    \begin{align*}
        \frac{1}{4}\norms{\bm P'_{j^*}\bm{\Delta}}\geq e^{-(\mu_{i^*}+\mu'_{j^*})s}\sum_{i\in \mathcal{I}}\norms{\bm P_i\ww},
    \end{align*}
    resulting in
    \begin{align*}
         \norms{\bm g(t)}&\geq \frac{3}{4}e^{-\mu'_{j^*}(T-t_0-s)}\norms{\bm P'_{j^*}\bm{\Delta}}=\frac{3}{4}e^{-m^*T}e^{\mu'_{j^*} s}\norms{\bm P'_{j^*}\bm{\Delta}}\\
         &\geq \frac{3}{4}e^{-m^*T}e^{\mu'_{j^*} S_+}\norms{\bm P'_{j^*}\bm{\Delta}}.
    \end{align*}
    After squaring using $\kappa^*=e^{-2m^* T}$, this gives
    \begin{align*}
        f_{\esp}(t_0+s)\geq \frac{9}{16}\norms{\bm P'_{j^*}\bm{\Delta}}^2e^{2\mu'_{j^*} S_+}\kappa^*
    \end{align*}
    for every integer $t_0+s\in[T]$ with $s\geq S_+$.

    \textbf{Putting everything together:}
    After decreasing $S_-$ and increasing $S_+$ if necessary, we have proved the following: there are $S_-<s^*-S_0-1$, $S_+>s^*+S_0+1$, both $S_\pm$ independent of $T$ and $T$ large enough (depending on both $S_-$ and $S_+$) such that
    \begin{itemize}
        \item according to \cref{step 5 discrete esp}, $f_{\esp}(\hat t_T)\leq 2C_1^2\kappa^*$;
        \item $f_{\esp}(t_0+s)\geq (81e^{-2\mu_{i^*} S_-}\smallnorm{\bm P_{i^*}\ww}_{\esp}^2/1024)\kappa^*$ for all $s\leq S_-$ with $t_0+s\in [T]$;
        \item $f_{\esp}(t_0+s)\geq (9\norms{\bm P'_{j^*}\bm{\Delta}}^2 e^{2\mu'_{j^*} S_+}/16)\kappa^*$ for all $s\geq S_+$ with $t_0+s\in [T]$.
    \end{itemize}
    Recall that $C_1$ is a constant independent of $S_{\pm}$ or $T$. We can accordingly increase $S_+$ and decrease (and make negative) $S_-$ such that
    \begin{align*}
        81e^{-2\mu_{i^*} S_-}\smallnorm{\bm P_{i^*}\ww}_{\esp}^2/1024\geq 3 C_1^2 \quad \text{and} \quad 9\norms{\bm P'_{j^*}\bm{\Delta}}^2 e^{2\mu'_{j^*} S_+}/16\geq 3C_1^2.
    \end{align*}
    So, for these choices of $S_-, S_+$ and $T$, we have that
    \begin{align}
        \mathcal T^*\subseteq[T]\cap[S_-+t_0,S_++t_0]. \label{step 5 discrete esp T star}
    \end{align}

    \bigskip
    \noindent\textbf{Step 6: Derivative sign in the central interval:} Recall that we are still working with the interpolator $f_{\esp}:[0,T]\rightarrow[0,\infty)$. Also, recall the definition of $S_0$ from Step 2. Now, decrease $S_-$ and increase $S_+$, if needed, such that $S_-<s^*-S_0-1<s^*+S_0+1<S_+$, and furthermore consider $R>0$ for which $[S_-,S_+]\subseteq[-R,R]$. Additionally, we can increase $T$ until $0<t_0-R<t_0+R<T$.
    
    By Step 3, we have
    \begin{align*}
        \underset{-R\leq s\leq R}{\sup}|\Xi(s)|\leq C_R e^{-\delta T}. 
    \end{align*}
    Note that, as $\Gamma(s)=0$ has a unique solution at $s^*$, on $[-R,R]\setminus [s^*-S_0,s^*+S_0]$ the function $\Gamma$ does not have a zero and moreover
    \begin{align*}
        {\min}\left\{|\Gamma(s)|: s\in[-R,s^*-S_0]\cup[s^*+S_0,R]\right\}>0,
    \end{align*}
    which is independent of $T$. So, we can take $T$ large enough so that
    \begin{align*}
        \underset{-R\leq s\leq R}{\sup}|\Xi(s)|\leq\frac{1}{2}{\min}\left\{|\Gamma(s)|: s\in[-R,s^*-S_0]\cup[s^*+S_0,R]\right\}.
    \end{align*}
    Accordingly, for every $s\in[-R,R]\setminus [s^*-S_0,s^*+S_0]$,
    \begin{align*}
        \text{sign}(\Xi(s)+\Gamma(s))=\text{sign}(\Gamma(s)),
    \end{align*}
    as $|\Xi|\leq |\Gamma|/2$ implies that addition of $\Xi$ cannot change the sign of $\Gamma$. But, as we showed in Step 2, $s^*$ is the unique root of $\Gamma$ and $\Gamma(s)<0$ for $s< s^*$, $\Gamma(s)>0$ for $s>s^*$, so there is no root of $\Xi+\Gamma$ in $[-R,R]\setminus [s^*-S_0,s^*+S_0]$.
    
    By Step 4, $z_T$ is the unique root of $\Gamma+\Xi$ in $[s^*-S_0,s^*+S_0]$, on which this function is strictly increasing. Together with the sign argument above, and recalling that $f'_{\esp}(t_0+s)=2\kappa^*(\Gamma(s)+\Xi(s))$ with $\kappa^*>0$, we obtain that for $\bar t_T:=t_0+z_T$,
    \begin{align*}
        f'_{\esp}(t)<0\quad\text{for } t\in[t_0+S_-,\bar t_T),
    \end{align*}
    and
    \begin{align*}
        f'_{\esp}(t)>0\quad\text{for } t\in(\bar t_T,t_0+S_+].
    \end{align*}
    Thus, on the central interval $[t_0+S_-,t_0+S_+]$ the interpolator $f_{\esp}$ is strictly decreasing before $\bar t_T$ and strictly increasing afterwards.
    
    \bigskip
    \noindent\textbf{Step 7: Localisation of the discrete global minimisers.} By Step 5 \cref{step 5 discrete esp T star}, every $t^*\in\mathcal{T}^*$ lies in $[t_0+S_-,t_0+S_+]$. From Step 6, the interpolator $f_{\esp}$ is strictly decreasing on $[t_0+S_-,\bar t_T]$, and strictly increasing on $[\bar t_T,t_0+S_+]$. By the choices of $S_\pm$ in Step 6 and the fact that $z_T\in[s^*-S_0,s^*+S_0]$, both $\lfloor \bar t_T\rfloor$ and $\lceil \bar t_T\rceil$ lie in the central interval $[t_0+S_-,t_0+S_+]$. Thus, the minimum of the discrete $f_{\esp}:[T]\rightarrow[0,\infty)$ over integer points in this interval can occur only at the two integers adjacent to $\bar t_T$, i.e.
    \begin{align*}
        \mathcal T^*\subseteq \{\lfloor \bar t_T\rfloor, \lceil \bar t_T\rceil\}.
    \end{align*}
    If $\bar t_T$ is an integer, these two points coincide. Hence, $\mathcal T^*$ contains at most two elements. Moreover, for every $t^*\in\mathcal T^*$, by the triangle inequality
    \begin{align*}
        |t^*-(t_0+s^*)|\leq |t^*-\bar t_T|+|\bar t_T-(t_0+s^*)|\leq 1+\frac{2C_{S_0}}{m_{S_0}}e^{-\delta T}.
    \end{align*}
    Absorbing $2C_{S_0}/m_{S_0}$ into $C$ and recalling $t_0=\mu_{j^*}' T/(\mu_{i^*}+\mu'_{j^*})$ gives the stated conclusion.
    \hfill$\blacksquare$

\subsection{Proof of \hyperref[prop:counterexample_nkh]{Proposition~\ref*{prop:counterexample_nkh}}}
We will work over $d=2$ dimensions. Define the following quantities:
\begin{itemize}
    \item Pretraining: take $n=2$, ridge parameter $\gamma=1/2$ and consider
    \begin{align*}
        \bm{X}=\begin{pmatrix}
            \sqrt{3} & 2/\sqrt{3}\\
            0 & \sqrt{5/3}
        \end{pmatrix},\quad \es=\frac{\bm X^\top\bm X}{n}=\begin{pmatrix}
            3/2 & 1\\
            1 & 3/2
        \end{pmatrix},\quad \bm H=\es+\gamma \bm I=\begin{pmatrix}
            2& 1\\ 1& 2
        \end{pmatrix}.
    \end{align*}
    Additionally, as $\bm{X}$ is invertible, we can select $\bm y$ appropriately such that
    \begin{align*}
        \ww=\frac{1}{n}\bm{H}^{-1}\bm X^\top \bm y=\begin{pmatrix}
            1 \\ 2
        \end{pmatrix}.
    \end{align*}
    (For example, for $\bm y=(8/\sqrt{3},14/\sqrt{15})^\top$).
    \item Fine-tuning: take $n'=2$, ridge parameter $\gamma'=1$ and consider
    \begin{align*}
        \bm X'=\begin{pmatrix}
            1 & 0\\
            1 & 0
        \end{pmatrix},\quad \esp =\frac{1}{n'} (\bm X')^\top \bm X'=\begin{pmatrix}
            1 & 0\\ 
            0 & 0
        \end{pmatrix}, \bm H'=\esp+\gamma'\bm I=\begin{pmatrix}
            2 & 0 \\
            0 & 1
        \end{pmatrix}.
    \end{align*}
    Additionally, for $\bm y'=(4,4)^\top$, we have
    \begin{align*}
        \ww'=\frac{1}{n'}(\bm H')^{-1} (\bm X')^\top \bm y'=\begin{pmatrix}
            2 \\ 0
        \end{pmatrix}.
    \end{align*}
    \item Thus, the empirical update is
    \begin{align*}
        \bm \Delta=\ww'-\ww=\begin{pmatrix}
            1 \\ -2
        \end{pmatrix}.
    \end{align*}
    \item We will take $\cp=\cf=1$ and $\eta=\eta'=1/4$.
\end{itemize}

Then,
\begin{align*}
    e^{-\bm G}=\bm I-\frac{1}{4}\bm H=\begin{pmatrix}
        1/2 & -1/4\\ -1/4 & 1/2
    \end{pmatrix},\quad e^{-\bm G}\ww=\begin{pmatrix}
        0 \\ 3/4
    \end{pmatrix}\in \operatorname{Ker}(\esp).
\end{align*}
Thus, (\ref{nkh}) fails at $t=1$. The burn-in defined in \hyperref[additionalnotation]{Appendix~\ref*{additionalnotation}} is $K=3$, since:
\begin{align*}
    \mu_1 = \ln(4/3),\quad \mu_2=\ln(4),\quad \frac{\norms{\bm P_2\ww}}{\norms{\bm P_1\ww}}=3,\quad\text{therefore } K=\left\lceil \frac{\ln(12)}{\ln(3)}\right\rceil = 3.
\end{align*}

Now, let's analyse $f_{\esp}(t)$. Recall that, due to the $\esp$-seminorm, $f_{\esp}(t)=g_1(t)^2$, where $g_1(t)$ is the first coordinate of
\begin{align*}
    \bm g(t)=(\bm I-e^{-(T-t)\bm G'})e^{-t\bm G}\ww-e^{-(T-t)\bm G'}\bm \Delta\ .
\end{align*}
For an integer $t$, let 
\begin{align*}
    \beta_t:=(e^{-t\bm G}\ww)_1=\frac{3-3^t}{2\cdot 4^t},
\end{align*}
so that
\begin{align*}
    g_1(t)=(1-2^{-(T-t)})\beta_t-2^{-(T-t)}.
\end{align*}
At $t:=1$, $\beta_1=0$ and $|g_1(1)|=2^{-(T-1)}$. For every $t\in\{2,\ldots,T\}$, $\beta_t<0$, so $|g_1(t)|\geq 2^{-(T-t)}>2^{-(T-1)}$, while at $t:=0$, $g_1(0)=1-2^{1-T}$, with magnitude greater than $2^{-(T-1)}$ for $T\geq 3$.

Hence, $t^*=1$ is the unique discrete minimiser for every $T\geq 3$, and $t^*/T\rightarrow 0$.
\hfill $\blacksquare$

\subsection{Proof of \hyperref[prop:random design]{Proposition~\ref*{prop:random design}}}
For fixed pretraining optimisation parameters, condition, if random, on the pretraining data $(\bm X, \bm y)$, so that $\bm G$ and $\ww$ are fixed. All subsequent probabilities are implicitly conditioned. Since $e^{-t\bm G}$ is invertible and $\ww\neq 0$, we have $e^{-t\bm G}\ww\neq 0$. Now, note that, writing the $i$th row of $\bm X'$ as $(\bm x_i')^\top$,
\begin{align*}
    \norms{e^{-t\bm G}\ww}^2=\frac{1}{n'}\sum_{i=1}^{n'}((\bm x_i')^\top e^{-t\bm G}\ww)^2,
\end{align*}
so that
\begin{align*}
    \{\norms{e^{-t\bm G}\ww}=0\}\subseteq \{(\bm x_1')^\top e^{-t\bm G}\ww=0\}.
\end{align*}
For each fixed integer $t\geq 0$, the set $\{\bm x'\in\mathbb{R}^d:(\bm x')^\top e^{-t\bm G}\ww=0\}$ is a hyperplane with Lebesgue measure $0$. Therefore,
\begin{align*}
    \mathbb{P}\left(\norms{e^{-t\bm G}\ww}=0\right)\leq \mathbb{P}\left((\bm x_1')^\top e^{-t\bm G}\ww=0\right)=0.
\end{align*}
Taking a countable union over all non-negative integers $t$, and averaging over the pretraining data if needed, gives the unconditional almost-sure conclusion.
\hfill$\blacksquare$

\subsection{Proof of \hyperref[prop:commutative]{Proposition~\ref*{prop:commutative}}}

Since $\es$ and $\esp$ commute, $\esp$ and $\bm{G}$ commute, so they can be simultaneously diagonalised by the same orthonormal basis $(\bm{u}_k)_{k=1}^d$, with corresponding eigenvalues $(g_k)_{k=1}^d$ for $\bm{G}$ and $(\sigma_k)_{k=1}^d$ for $\esp$ (here, we consider all eigenvalues). Let's write
\begin{align*}
    \ww=\sum_{k=1}^d a_k\bm{u}_k,
\end{align*}
so that
\begin{align}
\label{eq: norm ethww}
    e^{-t\bm{G}}\ww=\sum_{k=1}^d a_ke^{-tg_k}\bm{u}_k,\quad \norms{e^{-t\bm{G}}\ww}^2=\sum_{k=1}^d \sigma_ka_k^2e^{-2tg_k}.
\end{align}

(i) If $\ww\in\operatorname{ker}(\esp)$, then $a_k=0$ for all $1\leq k\leq d$ with $\sigma_k>0$, hence $\smallnorm{e^{-t\bm{G}}\ww}_{\esp}=0$.

To prove that $\mathcal{I}=\emptyset$, note that each projector $\bm{P}_i$ (for $1\leq i\leq r$) has the form $\bm{P}_i=\bm{U}\bm{D}_i \bm{U}^\top$, where $\bm{U}$ is the $d\times d$ matrix containing all eigenvectors $(\bm{u}_k)_{k=1}^d$ as columns, and $\bm{D}_i$ is a $d\times d$ diagonal matrix with $(\bm{D}_i)_{kk}=1$ whenever $\mu_i=g_k$, and $0$ otherwise. Note that $\ww\in\operatorname{ker}(\esp)$ implies that
\begin{align*}
    0=\bm{U}^\top (\esp\ww)=(\bm{U}^\top \esp \bm{U})(\bm{U}^\top \ww)=\operatorname{\textbf{diag}}(\sigma_1,\ldots,\sigma_d)(\bm{U}^\top \ww), 
\end{align*}
so that $\sigma_k(\bm{U}^\top \ww)_k=0$ for all $1\leq k\leq d$. Now, fix any $1\leq i\leq r$. Then
\begin{align*}
    \bm{U}^\top (\esp \bm{P}_i\ww)=(\bm{U}^\top \esp \bm{U})(\bm{U}^\top \bm{P}_i \bm{U})(\bm{U}^\top \ww)=\operatorname{\textbf{diag}}(\sigma_1,\ldots,\sigma_d)\bm{D}_i(\bm{U}^\top \ww).
\end{align*}
But, for each $1\leq k\leq d$, the $k$'th entry of $\operatorname{\textbf{diag}}(\sigma_1,\ldots,\sigma_d)\bm{D}_i(\bm{U}^\top \ww)$ is $\sigma_k(\bm{D}_i)_{kk}(\bm{U}^\top \ww)_k=0$, so $\bm{U}^\top (\esp \bm{P}_i\ww)=\bm{0}$. Because $\bm{U}$ is invertible we get  $\esp \bm{P}_i\ww=0$, hence $\bm{P}_i\ww\in\operatorname{ker}(\esp)$ and finally $\mathcal{I}=\emptyset$.

(ii) Assume $\ww\notin\operatorname{ker}(\esp)$. Then $\esp \ww=\sum_{k=1}^d a_k\sigma_k\bm{u}_k\neq 0$ and hence there exists $1\leq k\leq d$ with $\sigma_k>0$ and $a_k\neq 0$. So, the corresponding term in the summation in \cref{eq: norm ethww} is non-zero, and as $\sigma_k\geq 0$, $a_k^2\geq 0$, the whole sum strictly decreases with $t$. Hence, its minimum over $[K]$ is achieved at $t=K$. Finally, as $\mu_r$ is the largest eigenvalue of $\bm{G}$,
\begin{align*}
    \norms{e^{-K \bm{G}}\ww}^2=\sum_{k=1}^d\sigma_ka_k^2e^{-2K g_k}\geq e^{-2K \mu_r}\sum_{k=1}^d \sigma_ka_k^2=e^{-2K \mu_r}\norms{\ww}^2,
\end{align*}
which concludes the proof.\hfill $\blacksquare$

\section{Experimental details}
\label{app:experiments}
\paragraph{Data generation.} We generate 320 observations for each phase in $d=400$ dimensions. The rows of the random designs $\bm X,\bm X'$ are independently sampled from centred Gaussian distributions whose covariance matrices have the same eigenvalues, log-spaced from $1$ to $10^{-3}$, and independent eigenbases obtained by QR decompositions of matrices with entries sampled from standard normals. We draw independent $\bm w_\star,\bm \delta_\star\sim\mathcal N(\bm 0,\bm I_d/d)$ and generate the responses $\bm y, \bm y'$ as
\begin{align*}
    \bm y=\bm X \bm w_\star+\bm \varepsilon,\quad \bm y'=\bm X'(\bm w_\star+\bm \delta_\star)+\bm \varepsilon'
\end{align*}
with independent Gaussian noise entries in $\bm \varepsilon, \bm\varepsilon'$ of mean $0$ and variance $0.01$. Panels with 80 observations use the first 80 rows and corresponding labels from the corresponding dataset.

\paragraph{Training parameters.} All panels use ridge penalties $\gamma=\gamma'=0.05$, step-sizes $\eta=\eta'=0.05$, zero initialisation for pretraining, and the resulting checkpoint as the fine-tuning initialisation. The step-size conditions are checked numerically. We use costs $\cp=nd$, $\cf=n'd$ and the same training budget $B=65{,}536\times80d$, corresponding, under representation (\ref{eq:representation}), to $T={B}/\operatorname{lcm}(\cp,\cf)$. The training budget excludes estimation or any other computational overhead.

\paragraph{Predictions and numerical evaluation.} We use thin SVD to compute the ridge solutions and spectral quantities for the two-term prediction $\theta^* T+s^*$ (red dotted line). For the Lanczos estimation (grey dashed line) in \cref{subsec:complexity-separation}, the upstream bottleneck $\lambda_{i^*}$ is estimated using 40 matrix-free Lanczos iterations, initialised at $\bm X^\top \bm y/n$, with two-pass full reorthogonalisation. We estimate $\lambda_{i^*}$ as the smallest eigenvalue of the obtained tridiagonal matrix. The downstream bottleneck is $\lambda'_{j^*}=\gamma'$ in these instances, hence no downstream Lanczos iterations are needed. Every feasible integer allocation is evaluated to obtain the numerical optimum (black dot). The plots use logarithmic vertical axes with separate ranges. 

The experiment uses NumPy’s \texttt{default\_rng} with seed $2026$ and NumPy and SciPy in double-precision floating-point arithmetic. We use exponential rescaling to compute the closed-form GD residuals directly, and retain objective values in logarithmic form to avoid cancellation or underflow.

\end{document}